\documentclass[10pt,twocolumn,letterpaper]{article}

\usepackage[final,algorithms]{wacv}      

\usepackage{graphicx}
\usepackage{amsmath}
\usepackage{amssymb}
\usepackage{booktabs}

\usepackage[pagebackref,breaklinks,colorlinks]{hyperref}

\usepackage[capitalize]{cleveref}
\crefname{section}{Sec.}{Secs.}
\Crefname{section}{Section}{Sections}
\Crefname{table}{Table}{Tables}
\crefname{table}{Tab.}{Tabs.}

\usepackage[ruled]{algorithm2e}
\usepackage{cases}
\usepackage{amsfonts}
\usepackage{amsmath}
\usepackage{booktabs}
\usepackage{multirow}
\usepackage{xcolor}
\usepackage{tabularx}
\usepackage{caption}

\SetKwInput{KwData}{Input}
\SetKwInput{KwResult}{Output}

\SetKwProg{Fn}{Function}{}{}

\def\wacvPaperID{1623} 
\def\confName{WACV}
\def\confYear{2024}

\begin{document}

\title{Multi-View Person Matching and 3D Pose Estimation with Arbitrary Uncalibrated Camera Networks}

\author{Yan Xu\\
Carnegie Mellon University\\
{\tt\small yxu2@alumni.cmu.edu}
\and
Kris Kitani\\
Carnegie Mellon University\\
{\tt\small kmkitani@andrew.cmu.edu}
}
\maketitle

\begin{abstract}
Cross-view person matching and 3D human pose estimation in multi-camera networks are particularly difficult when the cameras are extrinsically uncalibrated.  Existing efforts generally require large amounts of 3D data for training neural networks or known camera poses for geometric constraints to solve the problem.  However, camera poses and 3D data annotation are usually expensive and not always available.  We present a method, PME, that solves the two tasks without requiring either information.  Our idea is to address cross-view person matching as a clustering problem using each person as a cluster center, then obtain correspondences from person matches, and estimate 3D human poses through multi-view triangulation and bundle adjustment.  We solve the clustering problem by introducing a ``size constraint" using the number of cameras and a ``source constraint" using the fact that two people from the same camera view should not match, to narrow the solution space to a small feasible region.  The 2D human poses used in clustering are obtained through a pre-trained 2D pose detector, so our method does not require expensive 3D training data for each new scene.  We extensively evaluate our method on three open datasets and two indoor and outdoor datasets collected using arbitrarily set cameras.  Our method outperforms other methods by a large margin on cross-view person matching, reaches SOTA performance on 3D human pose estimation without using either camera poses or 3D training data, and shows good generalization ability across five datasets of various environment settings.
\end{abstract}
\section{Introduction}
\label{sec:introduction}

\begin{figure}[t]
\begin{center}
\includegraphics*[clip=true,width=\linewidth]{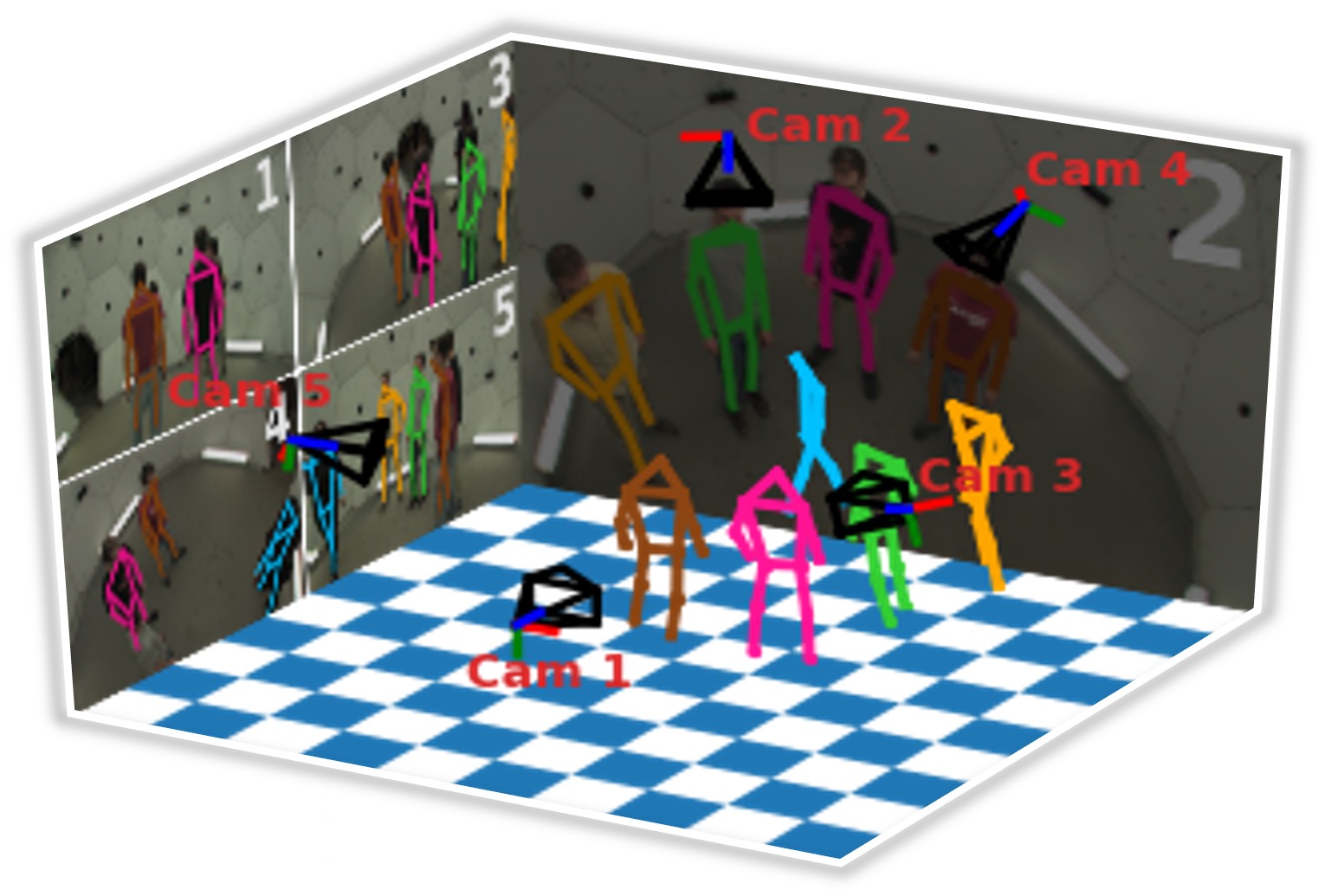}
\end{center}
\caption{Our result on CMU Panoptic studio dataset~\cite{joo2015panoptic}.  It jointly addresses cross-view person matching (colored) and 3D human pose estimation from multi-view image sequences captured by \textit{extrinsically uncalibrated} cameras. Camera poses (black) are also estimated.}
\label{fig:teaser}
\end{figure}

Matching people across multiple time-synchronized cameras and estimating their 3D pose are challenging, especially when camera poses are unknown.
Provided camera poses, existing efforts~\cite{dong2019fast,perez2022matching} often use the epipolar constraint to restrict the solution space of cross-view matching.
Works in 3D pose estimation~\cite{tu2020voxelpose,zhang2021direct,reddy2021tessetrack,wu2021graph} often use the camera poses to annotate 3D ground truth data for training pose regressors~\cite{lecun1998gradient,scarselli2008graph,vaswani2017attention}.
These methods rely heavily on calibrated camera poses, directly or indirectly. 
However, camera poses are not always available or expensive to obtain in many scenarios.  For example, can we perform 3D pose estimation for a group of people camping in a mountain or desert with multiple randomly placed cameras?

Aiming at answering the above question, we present a method for \textbf{P}e\textbf{r}s\textbf{o}n \textbf{M}atching and 3D pose \textbf{E}stimation (or \textbf{PME}) using image sequences captured from multiple camera views, \textit{without knowing the camera poses}.  Overall, our idea is to first match people in 2D across cameras, then obtain point correspondences from matched people, and finally solve and optimize 3D poses using the matched 2D poses.  This idea avoids the need for 3D data.  Meanwhile, since the matching step operates in 2D, it also can take full advantage of off-the-shelf 2D detection~\cite{cheng2020higherhrnet}, tracking~\cite{kuhn1955hungarian}, and re-identification (re-ID)~\cite{luo2019bag} models.  However, despite the benefits, the challenge also lies in 2D cross-view person matching under the condition of unknown camera poses.

To address cross-view person matching without camera poses, we carry on from two aspects: (1) Obtaining discriminative human feature representation, and (2) introducing a robust matching algorithm.  We obtain discriminative human feature representations by leveraging person re-ID and tracking.  Re-ID networks~\cite{solera2016tracking, zheng2015scalable} can encode similar appearance features for the same person and disparate features for different people.  However, relying only on re-ID features is not enough, especially when people have a similar appearance, \textit{e.g.}, clothes of the same color.  We thus use tracking~\cite{kuhn1955hungarian} and a ``max of sign voting" process to embed temporal information on top of the appearance feature.
Experiments show that such time-appearance feature encoding can provide discriminative human descriptions.

\begin{figure}[t]
\begin{center}
\includegraphics*[clip=true,width=\linewidth]{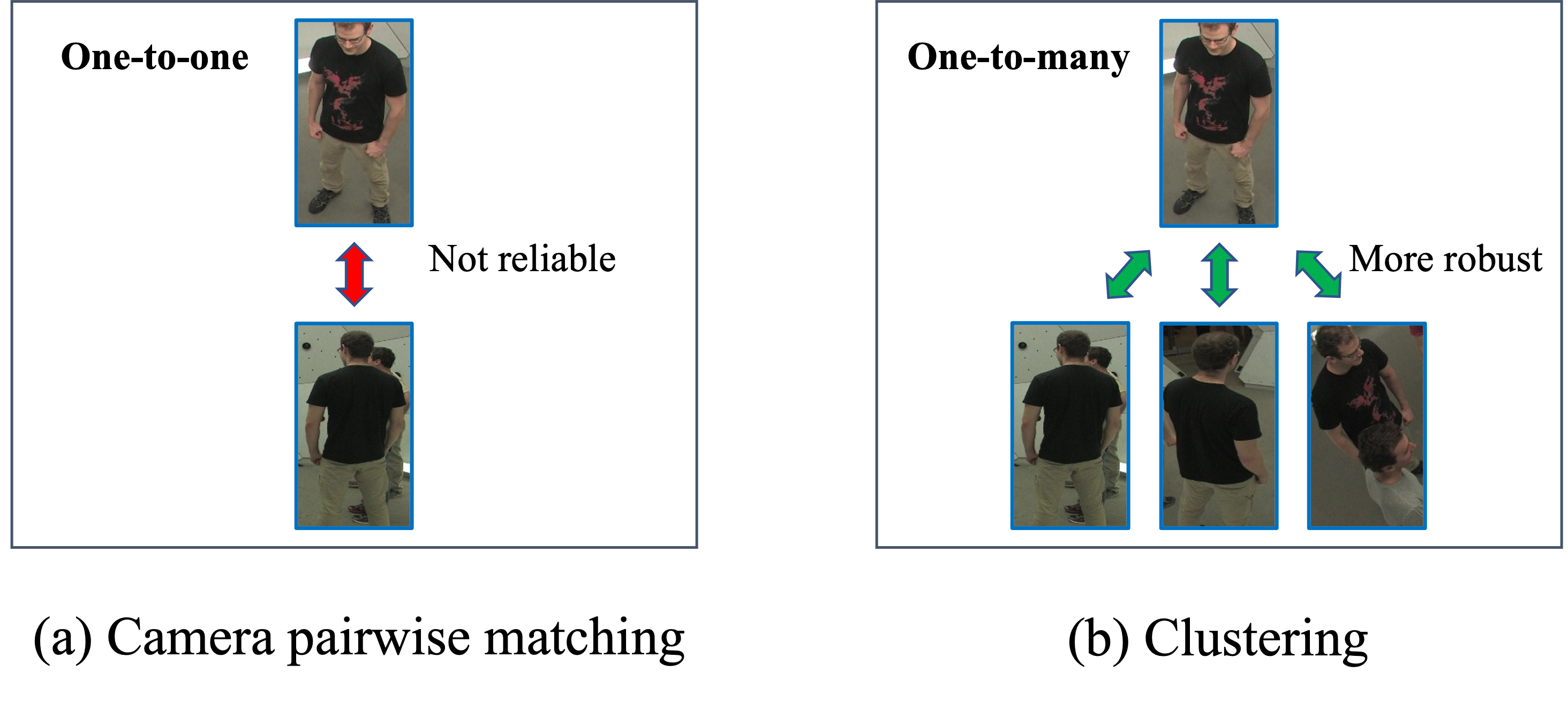}
\end{center}
\caption{Pair-wise matching \textit{vs} clustering. Pair-wise matching is less reliable, especially when the two camera angles are significantly different. Clustering is more robust by leveraging appearance information from multiple views.}
\label{fig:why_clustering}
\end{figure}

Aside from good feature representation, a robust cross-view matching algorithm is also required.  Existing methods~\cite{dong2019fast,perez2022matching} use epipolar cycle consistency computed from camera poses to restrict matching.  However, camera poses are unknown in our setting.  Without the strong epipolar constraint, one-to-one camera pairwise person matching would be unreliable. To address this, we propose to solve cross-view matching as clustering with each person as a cluster.  Compared to one-to-one camera pairwise matching, matching by clustering is more robust.  Fig.~\ref{fig:why_clustering} shows a comparison between camera pairwise matching and clustering.  As Fig.~\ref{fig:why_clustering} shows, when the camera pair observe the front and back of the person, the appearance feature from the two views could be significantly different, making one-to-one matching difficult.  As a comparison, clustering can still establish a correct match between the front and back of the person by using information from other cameras.



However, solely clustering is not enough.   We observe that solving matching as clustering without any constraints usually leads to local minima.  To address this, we introduce a ``size constraint" and a ``source constraint" to restrict the solution space to a small feasible region.  In particular, for a $N$-camera $K$-person system, ``size constraint" ensures the cluster size is smaller than $N$, and ``source constraint" guarantees that two people from the same camera view do not belong to the same cluster.  The two constraints can significantly narrow the solution space and make the global optimum much easier to reach.  However, according to \cite{wagstaff2001constrained}, the ``source constraint" is sensitive to the order of input data, meaning feeding data in a careless order will lead to incorrect clustering results.  We thus introduce a multi-step clustering algorithm, which first solves a size-constrained clustering, then detects false matches violating ``source constraint", and finally performs a source-constrained re-clustering on those false matches.  By leveraging the two constraints and the multi-step mechanism, PME significantly outperforms other clustering algorithms on cross-view human matching on three open datasets.

\begin{figure*}
\begin{center}
\includegraphics*[clip=true,width=0.99\textwidth]
{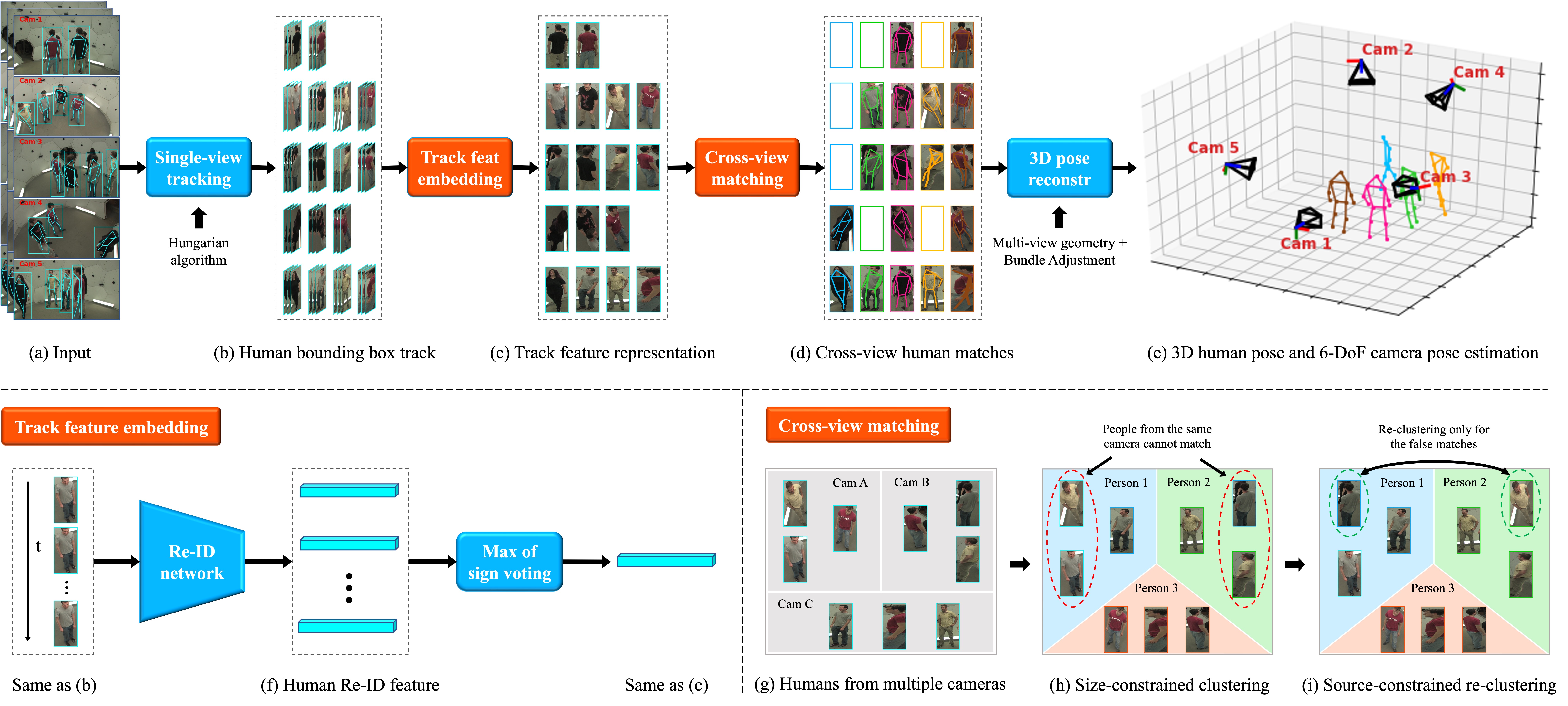}
\end{center}
\caption{Method overview.  Given \textbf{(a)} multi-view image sequences and 2D human pose detection, we first apply single-view tracking to obtain \textbf{(b)} human bounding box tracks.  We then embed \textbf{(c)} the track features using a pre-trained re-ID network and a "max of sign voting" process. With the multi-view features, we perform \textbf{(g-i)} a multi-step clustering algorithm to obtain \textbf{(d)} cross-view human matches.  Finally, we estimated \textbf{(e)} 3D human poses using multi-view geometry and bundle adjustment.}
\label{fig:system_overview}
\end{figure*}

After cross-view human matching, we associate corresponding body joints of the same person to obtain point correspondences, triangulate 3D human poses, and optimize the 3D poses across all camera views using bundle adjustment.  Fig.~\ref{fig:system_overview} presents an overview of our method.
We evaluate it on three open datasets, Shelf~\cite{belagiannis20143d}, Campus~\cite{belagiannis20143d}, and Panoptic~\cite{joo2015panoptic}, and two internal datasets, one indoor and one outdoor, collected using four extrinsically uncalibrated cameras.
On cross-view person matching, we compare with other clustering methods on the three open datasets.  Our method outperforms other methods on all datasets and leads by a large margin ($9.7\%$, $7.3\%$, $20.8\%$, $16.0\%$) on the most challenging Panoptic dataset~\cite{joo2015panoptic}.  On 3D human pose estimation, we compare with other methods, which require either camera poses or 3D data on Shelf and Campus~\cite{belagiannis20143d}.  PME reaches SOTA-level performance without requiring camera poses or 3D data.
Finally, we evaluate the generalization ability of PME on the two uncalibrated internal datasets by performing 3D pose estimation of people walking and interacting for 20 seconds (15 FPS).  It generalizes well without model tuning or parameter adjustment.


Our contributions are as follows:
(1) We present PME for jointly solving cross-view person matching and 3D pose estimation without the requirement of camera poses or 3D data;
(2) We introduce a ``size constraint", a ``source constraint", and a multi-step clustering algorithm to address cross-view person matching as constrained clustering when the epipolar constraint is unavailable;
(3) We introduce a ``max of sign voting" process to embed time-appearance feature from person track for discriminative human feature representation;
(4) We perform quantitative and qualitative evaluations on three open datasets and two internal datasets for a more comprehensive understanding of our method.

\section{Related work}
\label{sec:related_work}

Considering space limitation, we only discuss works in the following three most relevant research directions.

\paragraph{Cross-view human matching through clustering.}  Since our method is clustering-based, we compare it with other clustering algorithms. We use the same human feature representation for all algorithms.
The algorithms that we compare with include: KMeans-related methods~\cite{bradley2000constrained,wagstaff2001constrained,kanungo2002efficient, arthur2006k} that directly perform Expectation and Maximization~\cite{moon1996expectation}, Spectral method~\cite{ng2001spectral}, Hierarchical method~\cite{mullner2011modern}, DBSCAN~\cite{schubert2017dbscan}, and Gaussian Mixture Models~\cite{he2010laplacian}.  Size-constrained clustering was first used in \cite{bradley2000constrained}, and \cite{wagstaff2001constrained} proposed to convert background knowledge to ``must-link" and ``cannot-link" constraints.  Compared to them, our method first combines both constraints and adopts them in cross-view human matching.  Our method uses a multi-step clustering mechanism to address the data order sensitivity problem and outperforms all the baselines by a large margin.


\paragraph{Multi-view multi-person 3D human pose estimation.} Existing methods generally include two categories, multi-stage structural methods, and single-stage pose regressors.  Multi-stage structural methods~\cite{belagiannis2014multiple,belagiannis20153d,ershadi2018multiple,dong2019fast,perez2022matching} address the task in a multi-step fashion.  They first detect 2D human poses~\cite{li2019rethinking,sun2019deep,cao2017realtime,cheng2020higherhrnet}, then solve the 3D poses from the multi-view 2D poses.  Some structural methods~\cite{belagiannis20153d,ershadi2018multiple} solve multi-view matching and 3D pose estimation simultaneously by optimizing the 3DPS model~\cite{burenius20133d}, while others~\cite{dong2019fast,perez2022matching} solve multi-view matching first using appearance and geometry cues.  Our approach also solves multi-view matching first.  However, our approach does not require calibrated camera poses.   Single-stag pose regressors \cite{tu2020voxelpose,zhang2021direct,reddy2021tessetrack,zhang2022voxeltrack} directly regress 3D human poses from images/videos using deep networks.  These methods rely heavily on 3D training data, which need to be obtained using 2D pose annotation and calibrated camera poses.  Thus, 3D training data are generally very expensive to obtain, and annotating data for each new scene is impractical.  For these considerations, our method thus aims at avoiding the reliance on either calibrated camera poses or 3D data.


\paragraph{Person re-ID feature representation.}  We use person re-ID for human feature embedding in this work.  It is not our scope. Here, we briefly discuss some relevant works and explain our design choice.
Before the ethical problem came to attention, many efforts have been dedicated to addressing various challenges in re-ID, such as style variations caused by view changes\cite{zhong2018camera}, background clutter~\cite{li2018harmonious,song2018mask,si2018dual}, and human pose variations\cite{zhong2018camera,liu2018pose}.  Some works~\cite{kalayeh2018human,suh2018part} use both the global body feature and the local part feature to improve the performance.  In this work, we adopt a well-accepted strong re-ID baseline~\cite{luo2019bag} that uses a bag of tricks to embed human features.  We directly apply the  pre-trained network~\cite{solera2016tracking, zheng2015scalable} to our datasets without model tuning.

\section{Method}
\label{sec:method}

In this section, we elaborate details of our method shown in Fig.~\ref{fig:system_overview}.  Sec.~\ref{subsec:human_representation} introduces visual-temporal human feature embedding, which combines person-ID, single-view tracking, and a ``max-of-sign-voting" process.  Sec.~\ref{subsec:two_step_clustering} presents a multi-step clustering algorithm for solving cross-view person matching.  Sec.~\ref{subsec:3d_pose_est} expands on 3D human and camera pose estimation using the person matches.

\subsection{Visual-temporal human feature embedding}
\label{subsec:human_representation}

\subsubsection{Re-ID human feature representation}

Given multi-view image sequences, we first use an off-the-shelf detector~\cite{cheng2020higherhrnet} to detect the 2D bounding boxes and body joints.  We then adopt a re-ID network \cite{luo2019bag,xu2021wide,ma2022virtual} pre-trained on open datasets \cite{zheng2015scalable} to extract ``bottleneck layer" features from the human bounding boxes.  After that, we apply $L2$ normalization on the bounding box features to obtain the final re-ID feature representations for the people.

\subsubsection{Short-term tracking}

The re-ID features only contain appearance information, which is not discriminative enough, especially when people have similar appearances, \textit{e.g.}, similar cloth colors, hairstyles, \textit{etc}.  We thus use single-view tracking to introduce temporal information to the feature.  Consider a single-view video with $K$ people, let $\textbf{y}_t^k\in\mathbb{R}^D$ be the re-ID feature of person $k$ at time $t$, with $1\leq k \leq K$, features of all people at time $t$ and $t-1$ are: $\textbf{Y}_t = \{\textbf{y}_t^1, \textbf{y}_t^2, \cdots\}$ and $\textbf{Y}_{t-1} = \{\textbf{y}_{t-1}^1, \textbf{y}_{t-1}^2, \cdots\}$ (some people may be invisible to the cameras).  We use the Hungarian algorithm~\cite{kuhn1955hungarian} to establish one-to-one matches between $\textbf{Y}_t$ and $\textbf{Y}_{t-1}$:
\begin{equation}
\SetKwFunction{FHg}{Hunga}
    \{\textbf{y}_t^1, \textbf{y}_{t-1}^1\}, \{\textbf{y}_{t-1}^2, \textbf{y}_{t-1}^2\}, \cdots \gets \text{\FHg{$\textbf{Y}_t, \textbf{Y}_{t-1}$}}
    \label{eq:hungarian}
\end{equation}
We use cosine distance during matching and apply Eq.~\ref{eq:hungarian} between consecutive frames for $T$ time steps.  To minimize the impact of occlusion, we only apply short-term tracking with $T=10$.  Experiments show that tracking greatly improves feature discriminability and helps the matching.

\subsubsection{``Max of sign voting"}
\begin{figure}[t]
\centering
\includegraphics[width=0.96\linewidth]{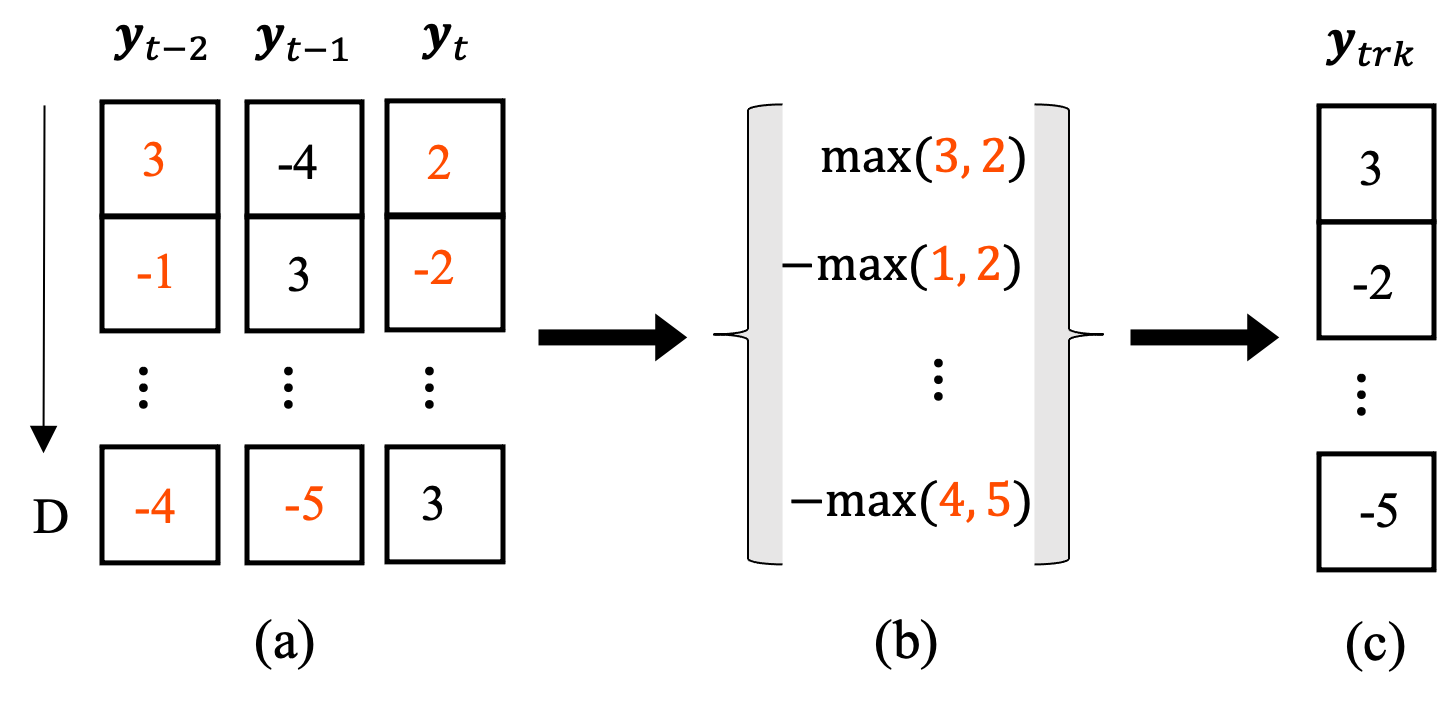}
\caption{Illustration of Max of Sign Voting.  \textbf{(a):} Track features $\{\textbf{y}_{t - 1}, \textbf{y}_{t - 1}, \textbf{y}_{t}\}$.  $D$ is the feature dimension.  In each row, the red elements are the ones with the dominant sign. \textbf{(b):} The process of ``max of sign voting".  \textbf{(c):} Feature representation of the track.}
\label{fig:max_of_sign_voting}
\end{figure}

With the associated track features $\{\textbf{y}_t,..., \textbf{y}_{t-T}\}$ (person id omitted), we aim to find $\textbf{y}_{trk}$ as the representation of the whole track.  Intuitively, since the tracking is short-term, the features should be close to each other, and directly taking the mean/max value as $\textbf{y}_{trk}$ seems reasonable.  However, this is not true.   Due to the over-sensitivity of the re-ID network, small pixel changes can cause dramatic differences between features, reflecting as sudden flips on the feature signs.  Directly taking the mean/max value with the sign flips leads to incorrect track feature representations.

We introduce a ``max of sign voting" process to deal with the sign flips.  The idea is to take the max value on features with the dominant sign and discard those incorrect minorities.  Formally, let $\textbf{Y}_{t, d} = \{y_{t, d}, \cdots, y_{t-T, d}\}$ be the tracking features of the $d$-th dimension.  We first find the dominant sign $s$ of $\textbf{Y}_{t, d}$ through voting.  After that, we take the max over absolute values of the elements with the dominant sign to obtain $|y_d|$.  Finally, we multiply $s$ with $|y_d|$ to get the track feature representation for the $d$-th dimension , $y_{trk, d}$.  We apply the process independently for each feature dimension. The process is formally defined as follows:
\begin{align}
\SetKwFunction{FVt}{SignVoting}
\SetKwFunction{FConcate}{Concate}
&s = \text{\FVt{$y_{t, d}, \cdots, y_{t-T, d}$}} \label{eq:sign_voting}\\
&|y_d| = \max(|y_{t, d}| * I(y_{t, d}, s), \cdots) \label{eq:max_abs}\\
&y_{trk, d} =  s * |y_d|  \label{eq:track_feat_d} \\
&\textbf{y}_{trk} =  \text{\FConcate{$y_{trk, 1}, \cdots, y_{trk, D}$}} \label{eq:track_feat}
\end{align}
In Eq.~\ref{eq:max_abs}, $I(y_{t, d}, s)=1$ if the sign of $y_{t, d}$ is $s$, else, $0$.
Fig.~\ref{fig:max_of_sign_voting} presents an illustration of ``max of sign voting".
Experiments show that tracking feature embedding with this process brings our method a notable improvement.  We also conduct experiments using max, mean, and ``mean of sign voting", and ``max of sign voting" shows a major lead.

\subsection{Multi-step clustering for cross-view matching}
\label{subsec:two_step_clustering}

Using the time-appearance human features as input, we now introduce a size- and source-constrained multi-step clustering algorithm for solving cross-view person matching.  The algorithm includes three steps.  Step 1 solves a size-constrained clustering problem and obtains an intermediate clustering result, $\mathcal{C}$.  Step 2 detects false matches in $\mathcal{C}$ that violate the source constraint.  Step 3 keeps the correct matches unchanged and performs a source-constrained post-clustering on the false-matched samples.  We detail each step in the following context.


\paragraph{Step 1: Size-constrained pre-clustering.}
Consider a $N$-camera$,K$-person system, let $\mathcal{D} = \{\textbf{x}_{i,j}\}$ be the 2D human feature representations, where $\textbf{x}_{i,j} \in \mathbb{R}^{D}$ is the feature of $j$-th people in the $i$-th camera view, $1 \leq i \leq N, 1 \leq j \leq K$.  Due to occlusion, each camera may only view a subset of the people.  We use $V_i$ to denote the number of people visible from camera $i$, and thus have $1 \leq V_i \leq K$, $1 \leq j \leq V_i$.  Our goal is to find $K$ cluster centers, $C^1, C^2, ..., C^{K}$, representing the $K$ people, such that the sum of the $L2$ distance between each sample $\textbf{x}_{i,j}$ and its nearest cluster center $C^{k}$ is minimized. The problem can be formulated as
\begin{align}
\min_{C, A} \quad & \sum_{i=1}^{N} \sum_{j=1}^{V_i} \sum_{k=1}^{K} A_{i, j}^{k} \cdot (\frac{1}{2} \| \textbf{x}_{i,j} - C^k \|_2^2) \label{eq:objective}\\
\textrm{s.t.} \quad & \sum_{k=1}^{K} A_{i, j}^{k} = 1, A_{i, j}^{k} \in \{0, 1\} \label{eq:sum_to_one}\\
              \quad &  \sum_{i=1}^{N}\sum_{j=1}^{V_i} W_{i, j}^{k} \geq 2 \label{eq:depth_ambiguity} \\
              \quad &  \sum_{i=1}^{N}\sum_{j=1}^{V_i} W_{i, j}^{k} \leq N \label{eq:size_constrained} \\
              \quad & 1 \leq i \leq N, 1 \leq j \leq V_i, 1 \leq k \leq K \label{eq:variable_range}
\end{align}
In Eq.~\ref{eq:sum_to_one}, $A_{i, j}^{k} = 1$ if $C^k$ is the closest cluster center for $\textbf{x}_{i,j}$, else $A_{i, j}^{k} = 0$.  Eq.~\ref{eq:depth_ambiguity} restricts a person to be observed by at least two cameras for solving the depth ambiguity.  Eq.~\ref{eq:size_constrained} is the size constraint ensuring the cluster size is no larger than the number of cameras $N$.

The above problem can be solved with fast network simplex algorithms following previous work~\cite{bradley2000constrained}.  The size-constrained clustering step can correctly match most humans and leaves a few hard samples for the following steps.

\paragraph{Step 2: False matches detection. } After obtaining the size-constrained clustering result, $\mathcal{C}$, our next step is to detect false matches violating the source constraint.  In particular, each human feature $\textbf{x}_{i,j}$ is associated with a camera label $n \in \{1, \cdots, N\}$.  We can identify samples from each cluster that have same camera label.  These samples are the false-matched samples that require a further process.

\paragraph{Step 3: Source-constrained post-clustering.} The last step is a source-constrained re-clustering on the false-matched samples, which always exist in the form of groups.  As an example, consider the following false matches: $\mathcal{V} = \{V_1, V_2\}$, where $V_1 = (\textbf{x}_{1, 1}, \textbf{x}_{1, 2})$ and $V_2 = (\textbf{x}_{2, 2}, \textbf{x}_{2, 3}, \textbf{x}_{2, 5})$.  $V_1$ means person 1 and 2 from camera 1 are incorrectly matched to one cluster, similar to person 2, 3, and 5 from camera 2 in $V_2$.  For convenience purposes, we omit the person id and camera id and denote $\mathcal{V}$ as $\mathcal{V} = \{\textbf{x}_1, \textbf{x}_2, \cdots\}$.  Our goal is to assign these false samples one by one to different clusters.  In each iteration, we first select the most distinguishable sample using a novel metric, \textit{{Sample Distinguishability Score (SDS).}\quad} Let the Euclidean distances between sample $\textbf{x}_{(\cdot)}$ and the cluster centers $\{C^1, ..., C^{K}\}$ be $\{d_1, ..., d_{K}\}$, we define \textit{SDS} as
\begin{numcases}{f(x)=}
   \frac{\min(d_1, ..., d_{K}; 2)}{\min(d_1, ..., d_{K}; 1)} & $n \geq 2$ \label{eq:sds_1}
   \\
   +\infty & $n = 1$ \label{eq:sds_2}
\end{numcases}
where $n\in\{1, \cdot, N\}$ is the number of clusters that do not contain people from the same camera view as $\textbf{x}_{(\cdot)}$, $\min(\cdot; n)$ denotes the $n$-th smallest value.  Eq.~\ref{eq:sds_1} measures the relative similarity between $\textbf{x}_{(\cdot)}$ and its closest and second closest cluster centers. Eq. (\ref{eq:sds_2}) means there is only one cluster that does not contain samples from the same camera view as $\textbf{x}_{(\cdot)}$.  
Larger $SDS$ indicates higher distinguishability.

\begin{figure}[t]
\centering
\includegraphics[width=\linewidth]{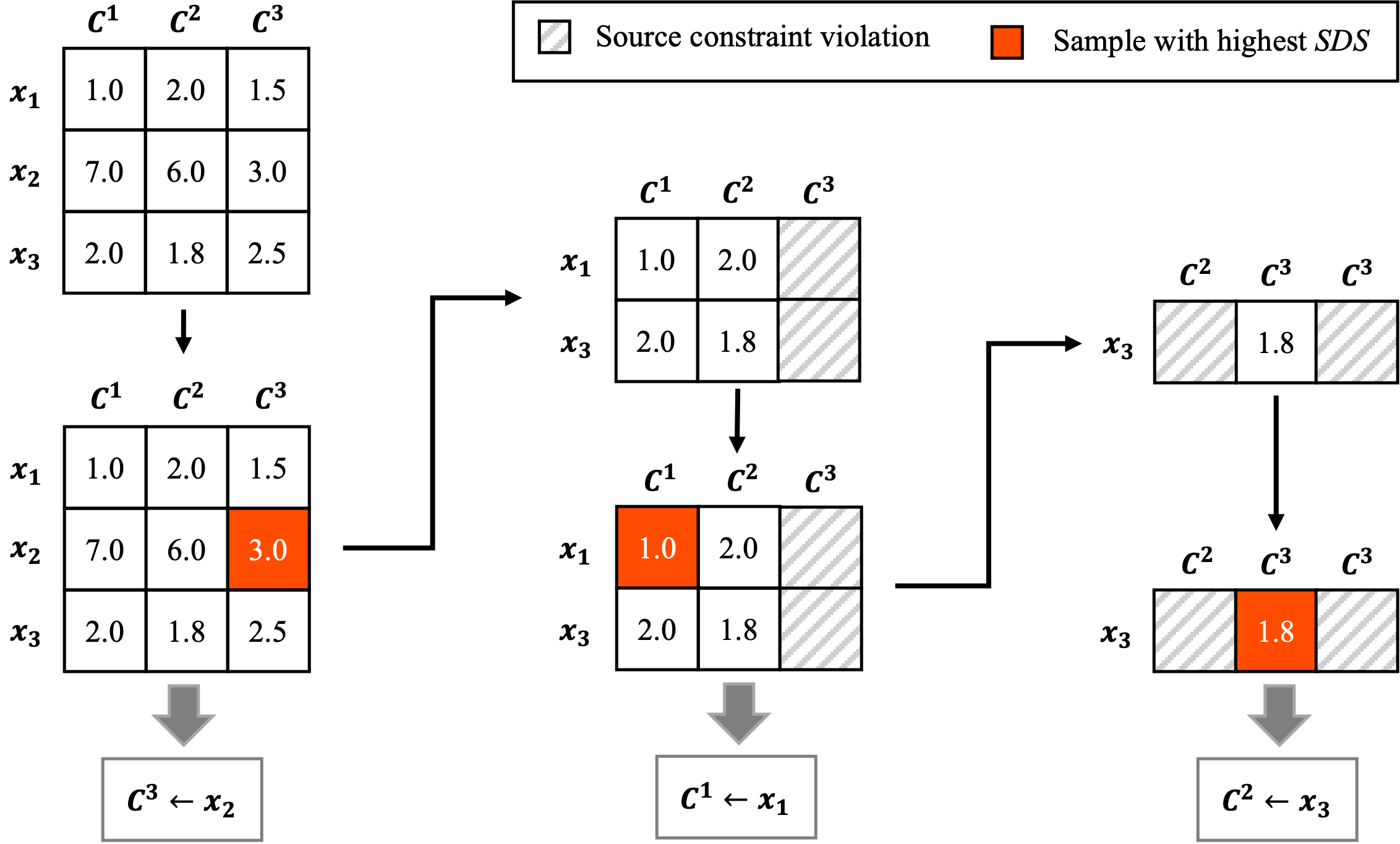}
\caption{Example of source-constrained post-clustering.  $\{x_\cdot\}$ are false-matched samples, $\{c^\cdot\}$ are clusters, numbers represents Euclidean distances.  Each iteration selects $x_\cdot$ with the highest \textit{SDS}.}
\label{fig:source_constrained_assign}
\end{figure}

In each iteration, we first sort the false-matched samples using their \textit{SDS} scores, then assign the sample with the highest \textit{SDS} to its closest clustering center, and remove the sample from $\mathcal{V}$ after the assignment.  We repeat this process until all the false-matched samples have been re-clustered.  Fig.\ref{fig:source_constrained_assign} illustrates Step 3 with an example.

The multi-step clustering mechanism can effectively leverage the source constraint without being impacted by the data order sensitivity problem~\cite{wagstaff2001constrained}.  We compare our multi-step clustering algorithm with other algorithms, our algorithm shows notable superiority.

\subsection{3D human pose estimation}
\label{subsec:3d_pose_est}

\paragraph{Cross-view point correspondences.}  With cross-view person matches, we can obtain point correspondences by associating the corresponding body joints across camera views, \textit{i.e.}, left hand to left hand, right knee to right knee, \textit{etc}. Facial detection is usually inaccurate since the organs, \textit{e.g.}, eyes, and ears, are small.  In experiments, we only use joints of the body and limbs.

\paragraph{Solving relative camera poses.}  We assume the intrinsic and distortion parameters are provided. Using the cross-view point correspondences, we first solve the relative poses between all camera pairs, by estimating the Essential matrix and decomposing it into a rotation and an up-to-scale translation. We then solve the scale ambiguity by assuming the length of the human lower leg to be $0.5m$.  After that, we align the estimated camera and 3D human to a common coordinate system using one camera as the world origin.

\paragraph{Bundle adjustment.} Finally, we globally optimize all camera and 3D human poses using bundle adjustment~\cite{triggs1999bundle}.  Since our method does not assume known camera poses, to evaluate it on standard human pose estimation datasets, we fix the camera poses to the ground truth during bundle adjustment.  Note that it is not necessary to have ground truth camera poses for our method.  For our uncalibrated data, we optimize both the camera and human poses.

\section{Experiment}
\label{sec:experiment}

We present evaluation results in this section, on both cross-view person matching and 3D human pose estimation.%
\begin{table*}[ht]
\renewcommand{\arraystretch}{1.}
\small
\centering
  \begin{tabular}{lcccccccc}
    \toprule
     & \multicolumn{4}{c}{Panoptic} & \multicolumn{4}{c}{Shelf} \\
     \cmidrule(r){2-9}
    \multicolumn{1}{l}{Method} & \multicolumn{1}{c}{Purity $\uparrow$} & \multicolumn{1}{c}{RI $\uparrow$} & \multicolumn{1}{c}{ARI $\uparrow$} & \multicolumn{1}{c}{F-Score $\uparrow$} & \multicolumn{1}{c}{Purity $\uparrow$} & \multicolumn{1}{c}{RI $\uparrow$} & \multicolumn{1}{c}{ARI $\uparrow$} & \multicolumn{1}{c}{F-Score $\uparrow$}\\
    \midrule
     Wagstaff \etal~\cite{wagstaff2001constrained} & $0.705$ & $0.794$ & $0.387$ & $0.517$ & $0.954$ & $0.945$ & $0.849$ & $0.925$\\
     Ng \etal~\cite{ng2001spectral} & $0.816$ & $0.865$ & $0.607$ & $0.693$ & $0.991$ & $0.985$ & $0.952$ & $0.980$\\
     Kanungo \etal~\cite{kanungo2002efficient} & $0.851$ & $0.892$ & $0.696$ & $0.765$ & $0.987$ & $0.982$ & $0.942$ & $0.976$\\
     Arthur \etal~\cite{arthur2006k} & $0.892$ & $0.921$ & $0.775$ & $0.826$ & $0.987$ & $0.982$ & $0.942$ & $0.976$\\
     He \etal~\cite{he2010laplacian} & $0.747$ & $0.815$ & $0.474$ & $0.593$ & $0.959$ & $0.945$ & $0.825$ & $0.935$\\
     Mullner \etal~\cite{mullner2011modern} & $0.889$ & $0.920$ & $0.770$ & $0.822$ & $0.986$ & $0.981$ & $0.938$ & $0.974$\\
     Schubert \etal~\cite{schubert2017dbscan} & $0.626$ & $0.722$ & $0.377$ & $0.551$ & $0.943$ & $0.874$ & $0.583$ & $0.809$\\
    \midrule
    Ours w/o tracking & $0.945$ & $0.969$ & $0.908$ & $0.927$ & $0.999$ & $0.999$ & $0.997$ & $0.998$\\
    Ours & $\textbf{0.989}$ & $\textbf{0.994}$ & $\textbf{0.983}$ & $\textbf{0.986}$ & $\textbf{0.999}$ & $\textbf{0.999}$ & $\textbf{0.998}$ & $\textbf{0.999}$\\
    \bottomrule
  \end{tabular}
\caption{Comparison with other clustering methods on cross-view person matching using Panoptic and Shelf datasets.}
\vspace{0pt}
\label{tab:clustering_panoptic_shelf}
\end{table*}
\begin{table*}[t]
\renewcommand{\arraystretch}{1.}
\centering
  \begin{tabular}{llcccccccc}
    \toprule
     & & \multicolumn{4}{c}{Campus} & \multicolumn{4}{c}{Shelf} \\
     \cmidrule(r){3-10}
     Require & \multicolumn{1}{l}{Method} & \multicolumn{1}{c}{Act. 1} & \multicolumn{1}{c}{Act. 2} & \multicolumn{1}{c}{Act. 3} & \multicolumn{1}{c}{Avg} & \multicolumn{1}{c}{Act. 1} & \multicolumn{1}{c}{Act. 2} & \multicolumn{1}{c}{Act. 3} & \multicolumn{1}{c}{Avg}\\
    \midrule
     \multirow{5}{*}{\shortstack[l]{3D Data\\(single-stage)}} & Huang \etal~\cite{huang2020end} & $98.0$ & $94.8$ & $97.4$ & $96.7$ & $98.8$ & $96.2$ & $97.2$ & $97.4$\\
     & Tu \etal~\cite{tu2020voxelpose}& $97.6$ & $93.8$ & $98.8$ & $96.7$ & $99.3$ & $94.1$ & $97.6$ & $97.0$\\
     & Zhang \etal~\cite{zhang2021direct} & $98.2$ & $94.1$ & $97.4$ & $96.6$ & $99.3$ & $95.1$ & $97.8$ & $97.4$\\
     & Huang \etal~\cite{huang2021dynamic} & $97.6$ & $93.7$ & $98.7$ & $96.7$ & $99.8$ & $\textcolor{blue}{96.5}$ & $97.6$ & $98.0$\\
     & Reddy \etal~\cite{reddy2021tessetrack} & $97.9$ & $\textcolor{blue}{95.2}$ & $99.1$ & $97.4$ & $99.1$ & $96.3$ & $98.3$ & $\textcolor{blue}{98.2}$\\
    \midrule
     \multirow{5}{*}{\shortstack[l]{Cam Pose\\(multi-stage)}} & Belagiannis \etal~\cite{belagiannis20153d} & $93.5$ & $75.7$ & $84.4$ & $84.5$ & $75.3$ & $69.7$ & $87.6$ & $77.5$\\
     & Ershadi \etal~\cite{ershadi2018multiple} & $94.2$ & $92.9$ & $84.6$ & $90.6$ & $93.3$ & $75.9$ & $94.8$ & $88.0$\\
     & Perez \etal~\cite{perez2022matching} & $98.4$ & $93.4$ & $98.3$ & $96.7$ & $98.9$ & $92.3$ & $97.8$ & $96.5$\\
     & Dong \etal~\cite{dong2019fast} & $97.6$ & $93.3$ & $98.0$ & $96.3$ & $98.8$ & $94.1$ & $97.8$ & $96.9$\\
     & Dong \etal~\cite{dong2019fast} \textit{w/o cam pose} & $\textit{97.6}$ & $\textit{93.3}$ & $\textit{96.5}$ & $\textit{95.8}$ & $\textit{98.6}$ & $\textit{60.5}$ & $\textit{94.3}$ & $\textit{84.5}$\\
    \midrule
    \shortstack[l]{Neither} & Ours & $\textbf{99.0}$ & $\textbf{94.7}$ & $\textbf{99.6}$ & $\textbf{97.8}$ & $\textbf{99.6}$ & $\textbf{95.2}$ & $\textbf{98.5}$ & $\textbf{97.8}$\\
    \bottomrule
  \end{tabular}
\caption{Comparison to other 3D human pose estimation methods on Campus and Shelf datasets.  Our method reaches state-of-the-art comparable performance and is the only method that does not require either 3D training data or calibrated camera poses.}
\label{tab:human_pose_estimation}
\end{table*}
\begin{table}[t]
\renewcommand{\arraystretch}{1.}
\small
\centering
  \begin{tabularx}{\columnwidth}{lcccc}
    \toprule
    \multicolumn{1}{l}{Method} & \multicolumn{1}{c}{Purity $\uparrow$} & \multicolumn{1}{c}{RI $\uparrow$} & \multicolumn{1}{c}{ARI $\uparrow$} & \multicolumn{1}{c}{F-Score $\uparrow$}\\
    \midrule
     Wagstaff \etal~\cite{wagstaff2001constrained} & $0.957$ & $0.949$ & $0.673$ & $0.919$\\
     Ng \etal~\cite{ng2001spectral} & $0.998$ & $0.995$ & $0.981$ & $0.995$\\
     Kanungo \etal~\cite{kanungo2002efficient} & $0.998$ & $0.998$ & $0.994$ & $0.998$\\
     Arthur \etal~\cite{arthur2006k} & $0.998$ & $0.998$ & $0.994$ & $0.998$\\
     He \etal~\cite{he2010laplacian} & $0.975$ & $0.963$ & $0.832$ & $0.949$\\
     Mullner \etal~\cite{mullner2011modern} & $0.998$ & $0.998$ & $0.994$ & $0.998$\\
     Schubert \etal~\cite{schubert2017dbscan} & $0.988$ & $0.958$ & $0.799$ & $0.933$\\
    \midrule
    Ours w/o tracking & ${1.000}$ & ${1.000}$ & ${1.000}$ & ${1.000}$\\
    Ours & $\textbf{1.000}$ & $\textbf{1.000}$ & $\textbf{1.000}$ & $\textbf{1.000}$\\
    \bottomrule
  \end{tabularx}
\vspace{0pt}
\caption{Comparison with other clustering methods on cross-view person matching using Campus dataset.}
\label{tab:clustering_campus}
\end{table}

\subsection{Datasets}
\label{subsec:exp_dataset}

\paragraph{Open datasets.}  Following previous works, we evaluate our method on three open datasets: CMU Panoptic~\cite{joo2015panoptic}, Shelf and Campus~\cite{belagiannis20143d}.  Panoptic captures people doing various activities in an indoor studio.  Campus captures three people walking in an outdoor environment.  Shelf captures four people disassembling a shelf.  All three datasets provide ground truth camera poses and 3D human poses.  Note that we use one of the cameras as the world origin in our method. For fair comparisons, we first align our results to the coordinate system defined by the datasets, and then compare with other methods.

\paragraph{Data from arbitrary uncalibrated camera networks.}  We collect two datasets using four extrinsically uncalibrated cameras to evaluate the generalization ability of our method.  For convenience, we name the two datasets Office and Square.  Office is captured in a $6m\times8m$ indoor environment, while Square is captured in a  $15m\times20m$ outdoor environment.  Both datasets capture three people walking and interacting and do not provide camera and human pose annotation.  The intrinsic parameters are provided.

\subsection{Cross-view person matching}
\label{subsec:exp_human_matching}

\paragraph{Baselines and Metrics.} We compare our method with other clustering algorithms on cross-view person matching.  The baselines include KMeans-related methods~\cite{wagstaff2001constrained,kanungo2002efficient, arthur2006k}, Spectral method~\cite{ng2001spectral}, Hierarchical method~\cite{mullner2011modern}, DBSCAN~\cite{schubert2017dbscan}, and Gaussian Mixture Models~\cite{he2010laplacian}.
All methods use the same human feature representations.  Following previous work~\cite{palacio2019evaluation}, we report four standard metrics: Purity, Rand Index (RI), Adjusted Rand Index (ARI), and F-Score.

\paragraph{Comparison with baselines.} Tab.~\ref{tab:clustering_panoptic_shelf} and Tab.~\ref{tab:clustering_campus} present the comparison between our multi-step source and sized constrained method and other clustering algorithms.  From the results, we have the following observations: 1. Our method outperforms all the baselines across all three datasets.  2. Our method leads other methods by a large margin on the challenging Panoptic and Shelf datasets.  It is very difficult to perform person matching for Shelf and Panoptic datasets since they have severe occlusion with multiple people interacting in a small space.  The result shows that the robustness of our method against occlusion.  3. Our method also reaches the best performance on Campus dataset even without using the temporal information through single-view tracking.  Occlusion is light in Campus dataset since people are mostly standing static, so other methods also perform reasonably well.  However, from the performance change from to Campus, to Shelf, then to Panoptic, we can peek the robustness of our method compared with other methods.

\subsection{3D human pose estimation}
\label{subsec:exp_pose_estimation}

\paragraph{Baselines and Metrics.}   Existing two categories of 3D human pose estimation methods require either 3D training data~\cite{huang2020end,tu2020voxelpose,zhang2021direct,huang2021dynamic,reddy2021tessetrack} or calibrated camera poses~\cite{belagiannis20153d,ershadi2018multiple,dong2019fast,perez2022matching}.  We compare our method with them and report Percentage of Correct Parts (PCP) following previous works.

\paragraph{Comparison with baselines.}  Tab.\ref{tab:human_pose_estimation} presents comparisons between our method and the baseline methods on Shelf and Campus datasets. From the results, we have the following observations: 1. Our method, without using camera poses, outperforms all of the multi-stage methods that require camera poses as prior knowledge.  2. We want to specifically mention one of the most impactful multi-stage methods in recent years, MVPose~(\cite{dong2019fast}), since it can also function without using camera poses. However, its performance drops dramatically without using camera poses.  As a comparison, our method is more stable and outperforms MVPose~(\cite{dong2019fast}) even when it uses camera poses.  3. Our method outperforms or is comparable with all the single-stage methods. These methods require labeled 3D data and model training specifically for each new scene/dataset.  Compared with them, our method has much better generalization ability without requiring specific 3D data or model training.

\begin{figure*}[t]
\centering
\includegraphics[width=0.96\linewidth]{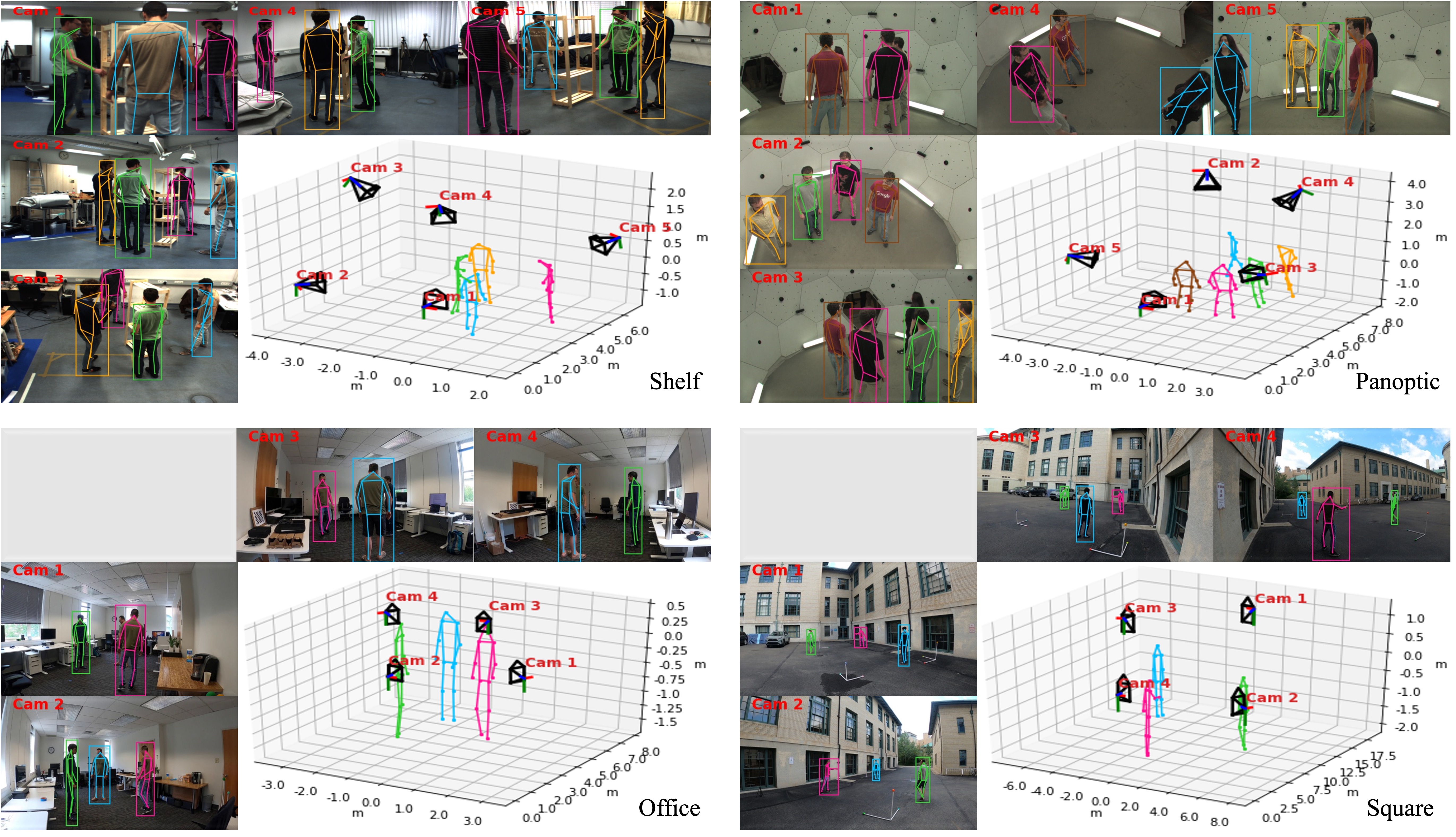}
\caption{Qualitative result of our method on cross-view person matching, 3D human pose estimation, as well as camera pose estimation. We present results on two open datasets (top row) and two uncalibrated internal datasets (bottom row).  People are matched by colors 2D-2D and 2D-3D.  Estimated camera poses are represented by black rigid bodies.  (Please consider zooming in for a better visual effect.)}
\label{fig:3d_wild_pose}
\end{figure*}

\paragraph{Qualitative visualization.}  Fig.\ref{fig:3d_wild_pose} presents qualitative visualization of our method on two open datasets and two datasets (indoor and outdoor) collected using arbitrary uncalibrated cameras.  The result shows the generalization ability of our method.  We present more results in the supplementary materials, please refer to it for more details.

\subsection{Ablation study}
\label{subsec:exp_ablation}

\paragraph{Temporal information.}  Tab.~\ref{tab:clustering_panoptic_shelf} presents ablation study results on the impact of temporal information introduced by single-view tracking.  As Tab.~\ref{tab:clustering_panoptic_shelf} shows, introducing temporal information through tracking greatly boosts performance, especially when occlusion is severe. On the most challenging Panoptic dataset, our method leads by $\textbf{9.7\%$, $7.3\%$, $20.8\%$, $16.0\%}$.
\begin{table}[t]
\renewcommand{\arraystretch}{1.}
\small
\centering
  \begin{tabularx}{\columnwidth}{lcccc}
    \toprule
    \multicolumn{1}{l}{Method} & \multicolumn{1}{c}{Purity $\uparrow$} & \multicolumn{1}{c}{RI $\uparrow$} & \multicolumn{1}{c}{ARI $\uparrow$} & \multicolumn{1}{c}{F-Score $\uparrow$}\\
    \midrule
     Mean & $0.947$ & $0.970$ & $0.911$ & $0.930$\\
     Max & $0.973$ & $0.985$ & $0.954$ & $0.979$\\
     Mean of sign voting & $0.948$ & $0.970$ & $0.912$ & $0.931$\\
     \midrule
    Max of sign voting & $\textbf{0.989}$ & $\textbf{0.994}$ & $\textbf{0.983}$ & $\textbf{0.986}$\\
    \bottomrule
  \end{tabularx}
\vspace{0pt}
\caption{Ablation study on temporal information embedding. ``Max of sign voting" outperforms all other variants.}
\label{tab:max_of_sign_voting}
\end{table}
\begin{table}[t]
\renewcommand{\arraystretch}{1.}
\small
\begin{center}
  \begin{tabular}{ccccccc}
    \toprule
     & \multicolumn{3}{c}{Campus} & \multicolumn{3}{c}{Shelf} \\
     \cmidrule(r){2-7}
     \multirow{-2}{*}{Noise} & \multicolumn{1}{c}{A1} & \multicolumn{1}{c}{A2} & \multicolumn{1}{c}{A3} & \multicolumn{1}{c}{A1} & \multicolumn{1}{c}{A2} & \multicolumn{1}{c}{A3}\\
     \midrule
     No extra noise & ${99.0}$ & ${94.7}$ & ${99.6}$ & ${99.6}$ & $\textbf{95.2}$ & $\textbf{98.5}$\\
     \midrule
     $2\times2$ & $99.0$ & $94.7$ & $98.6$ & $99.6$ & $95.2$ & $98.5$\\
     $4\times4$ & $99.0$ & $92.4$ & $98.4$ & $99.6$ & $93.6$ & $97.8$\\
     $6\times6$ & $97.3$ & $87.6$ & $95.2$ & $96.2$ & $88.5$ & $95.7$\\
     $10\times10$ & $87.3$ & $82.7$ & $85.1$ & $88.0$ & $81.4$ & $94.1$\\
     $20\times20$ & $58.5$ & $61.5$ & $62.4$ & $76.3$ & $58.6$ & $82.0$\\
    \bottomrule
  \end{tabular}
\end{center}
\vspace{-10pt}
\caption{Ablation study on detection noise robustness. $N\times N$ means uniformly sampling a position from a $N\times N$ pixel square area centered at the detected joint to simulate noise.}
\label{tab:noise_robustness}
\end{table}

\paragraph{Max of sign voting.}  Tab.\ref{tab:max_of_sign_voting} presents ablation study results on ``max of sign voting".  We use the most challenging Panoptic dataset in this experiment for a more obvious result and better understanding.  As Tab.\ref{tab:max_of_sign_voting} shows, compared to directly taking the mean or max value and a ``mean of sign voting" variant, our ``max of sign voting" performs the best.

\paragraph{Analysis on detection noise robustness.}  Since our method is multi-stage, the quality of 2D pose detection in the first stage plays an important role.  We conducted this experiment to analyze the detection noise robustness of our method and present the result in Tab.\ref{tab:max_of_sign_voting}. As the result shows, our method can perform reasonably well with $6\times6$ detection noise.  Since the noise is independently added to each body joint, the noise added to different body joints could be various in magnitude and direction, so $6\times6$ noise is considerably large.  The 2D pose detector that we use~\cite{cheng2020higherhrnet} in this paper can provide decent enough 2D pose detection without the need for any fine-tuning.

\section{Conclusion}
\label{sec:conclusion}

We presented PME, a method for jointly addressing multi-view person matching and 3D pose estimation.  We showed that PME can work well without the need for calibrated camera poses or 3D training data.
We introduced a multi-step clustering algorithm to solve cross-view person matching as a constrained clustering problem and a ``max-of-sign-voting" mechanism to embed temporal information into human features for a more representative and discriminative representation.  We performed quantitative and qualitative evaluations of our method on both open datasets and two indoor and outdoor datasets collected using arbitrary uncalibrated cameras.  Our method significantly outperforms other methods on cross-view matching and reaches state-of-the-art performance on 3D human pose estimation without either 3D data or camera pose information.



%
\begin{table*}[ht]
\begin{center}
\begin{tabular}{ c  p{5cm}  p{5cm}  }
    \toprule
    Dataset \& Result & Setting & Link to Video \\
    \cmidrule(r){1-1}\cmidrule(lr){2-2}\cmidrule(l){3-3}
    \raisebox{-\totalheight}{\includegraphics[width=0.3\textwidth]{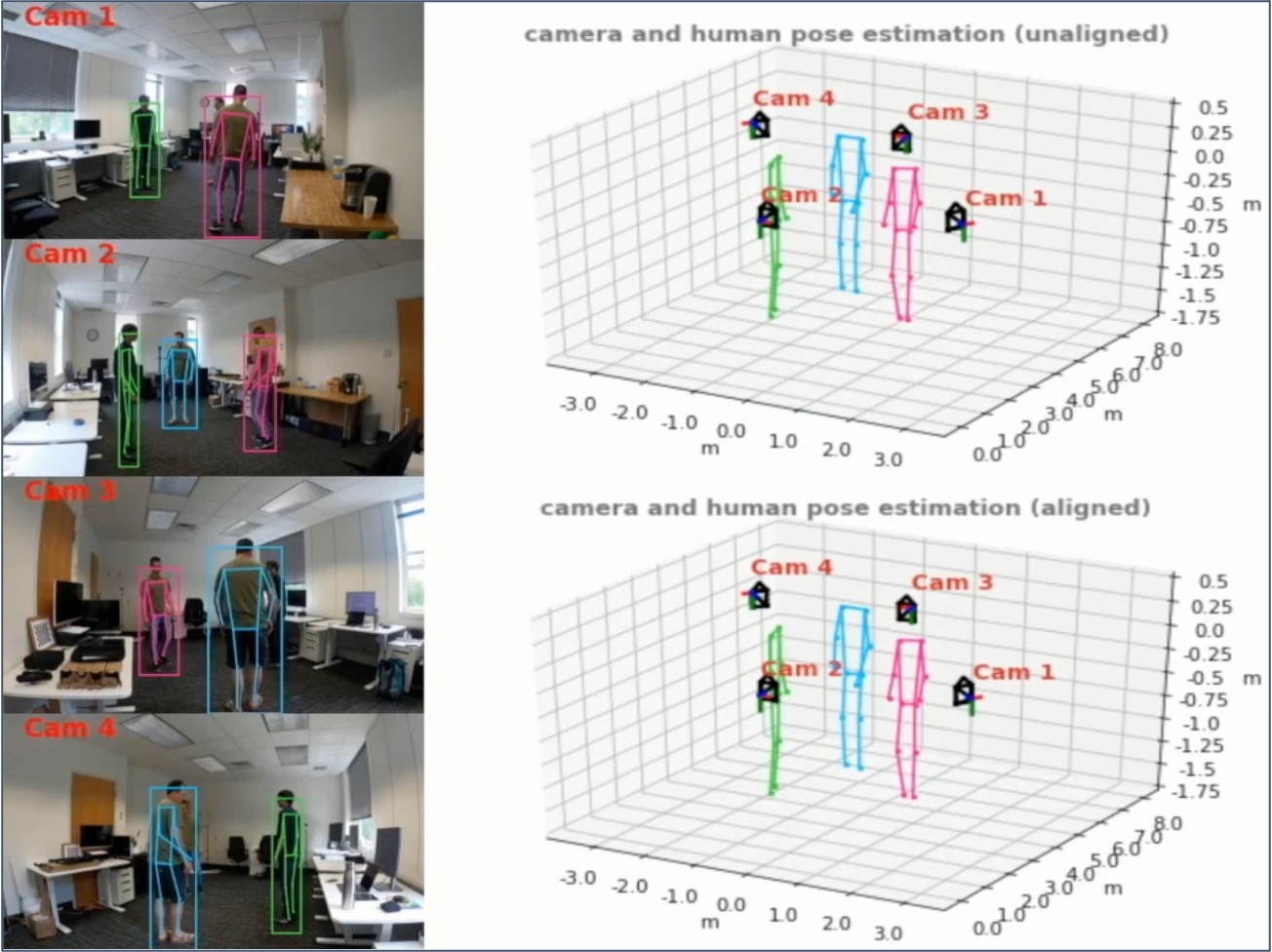}}
    & 
    \begin{itemize}
    \item Camera poses: Unknown
    \item Environment: Indoor
    \item Scene size: $\sim 6m\times 8m$
    \item \#Cameras: 4
    \item \#People: 3
    \end{itemize}
    & 
    \begin{itemize}
    \item \href{https://drive.google.com/file/d/1KA3oE6gueod0i-7YlY8n6qc8Vl9yWSzW/view?usp=sharing}{\textcolor{teal}{Video\_Result\_Office}} (collected with arbitrary indoor cameras)
    \end{itemize} \\
    
    \cmidrule(r){1-1}\cmidrule(lr){2-2}\cmidrule(l){3-3}
    \raisebox{-\totalheight}{\includegraphics[width=0.3\textwidth]{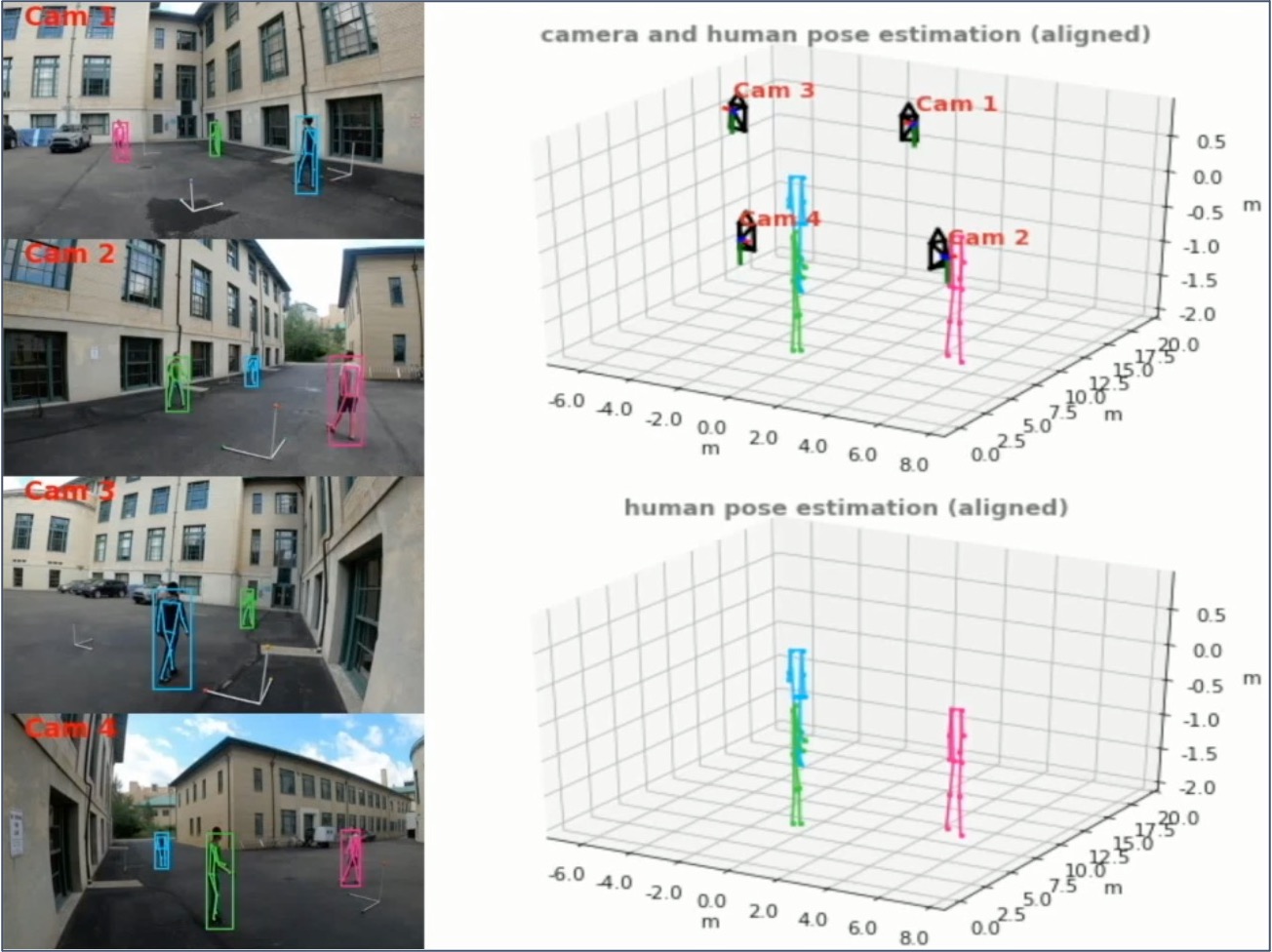}}
    & 
    \begin{itemize}
    \item Camera poses: Unknown
    \item Environment: Outdoor
    \item Scene size: $\sim 16m\times 28m$
    \item \#Cameras: 4
    \item \#People: 3
    \end{itemize}
    & 
    \begin{itemize}
    \item \href{https://drive.google.com/file/d/1oWjp0lfVTCTiAXVGlV1hFTSU4N9O0bm3/view?usp=sharing}{\textcolor{teal}{Video\_Result\_Square}} (collected with arbitrary outdoor cameras)
    \end{itemize} \\
    
    \cmidrule(r){1-1}\cmidrule(lr){2-2}\cmidrule(l){3-3}
    \raisebox{-\totalheight}{\includegraphics[width=0.3\textwidth]{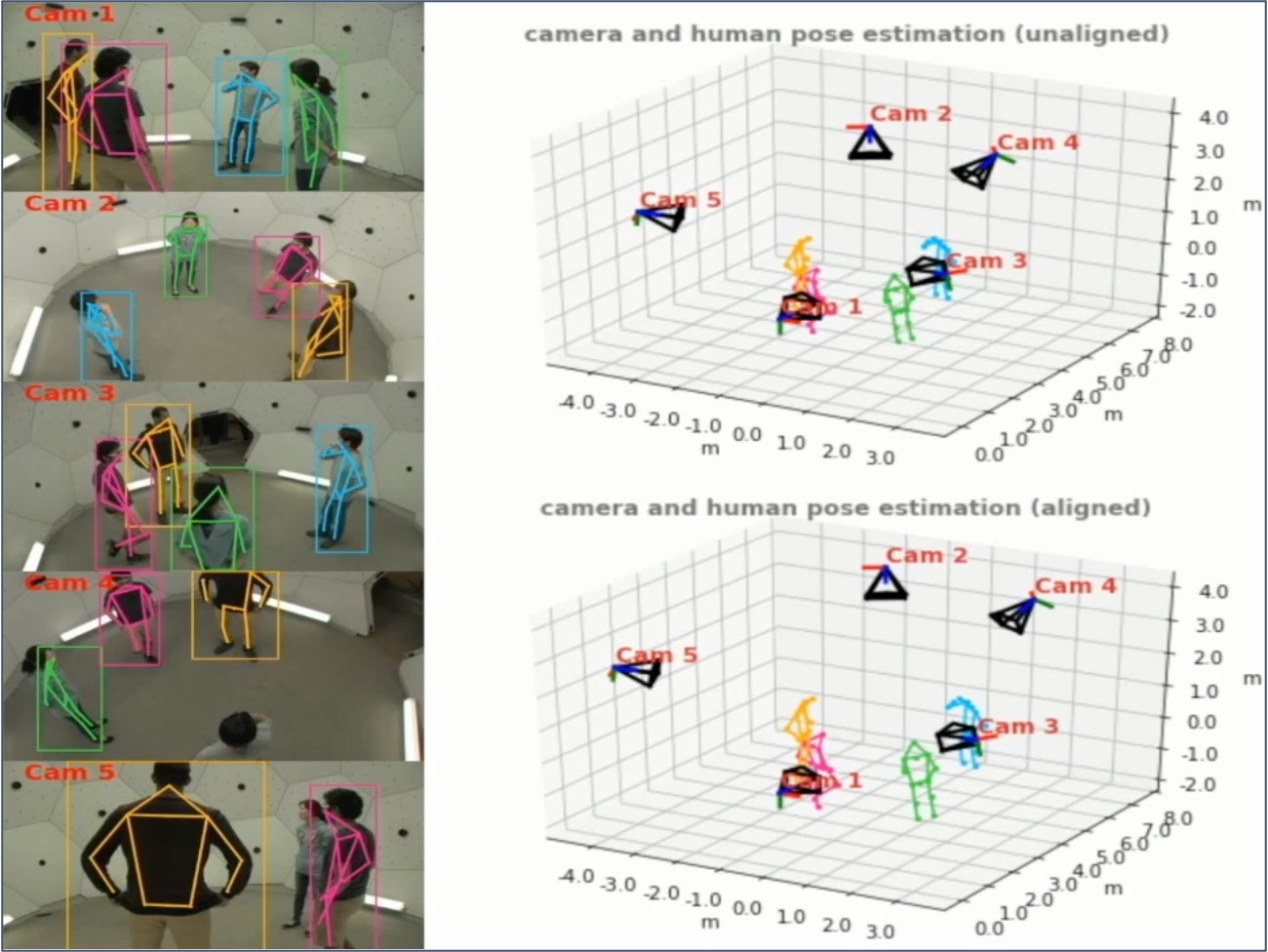}}
    & 
    \begin{itemize}
    \item Camera poses: Unknown
    \item Environment: Indoor
    \item Scene size: $\sim 8m\times 8m$
    \item \#Cameras: 5
    \item \#People: 3 - 4
    \end{itemize}
    & 
    \begin{itemize}
    \item \href{https://drive.google.com/file/d/1wqPpXJ0Egwll-GaIHwJBw1PkZ7TIu7gA/view?usp=sharing}{\textcolor{teal}{Video\_Result\_Panoptic\_Seq1}}
    \item \href{https://drive.google.com/file/d/11G3QYmCVU7gjXCCT-W200oLyB3QPCBPb/view?usp=sharing}{\textcolor{teal}{Video\_Result\_Panoptic\_Seq2}}
    \end{itemize} \\
    
    \cmidrule(r){1-1}\cmidrule(lr){2-2}\cmidrule(l){3-3}
    \raisebox{-\totalheight}{\includegraphics[width=0.3\textwidth]{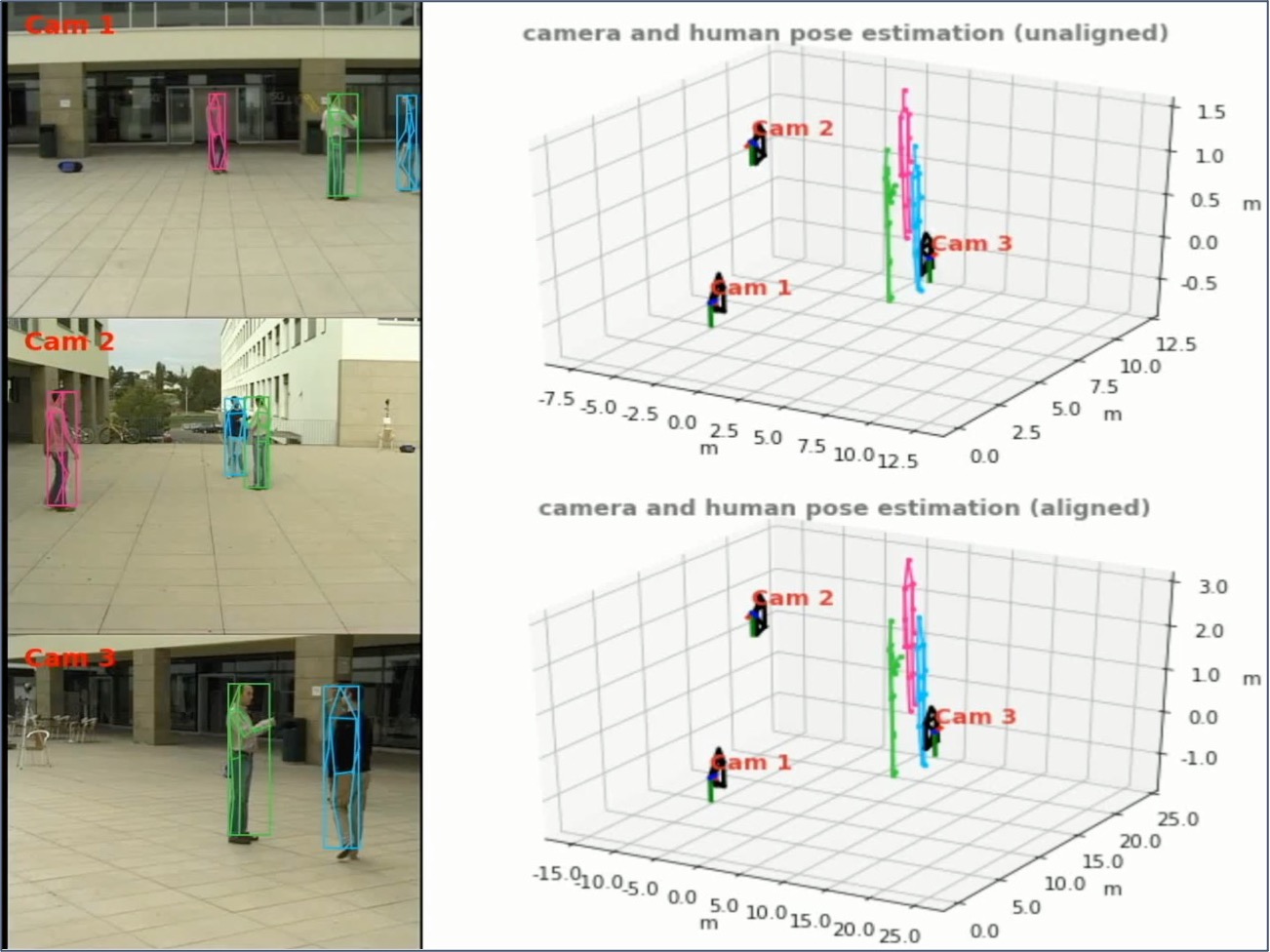}}
    & 
    \begin{itemize}
    \item Camera poses: Unknown
    \item Environment: Outdoor
    \item Scene size: $\sim 12m\times 20m$
    \item \#Cameras: 3
    \item \#People: 3
    \end{itemize}
    & 
    \begin{itemize}
    \item \href{https://drive.google.com/file/d/1cE8BUC5fH17kzAwvHw-pWBBxAEseu-kY/view?usp=sharing}{\textcolor{teal}{Video\_Result\_Campus}}
    \end{itemize} \\
    
    \cmidrule(r){1-1}\cmidrule(lr){2-2}\cmidrule(l){3-3}
    \raisebox{-\totalheight}{\includegraphics[width=0.3\textwidth]{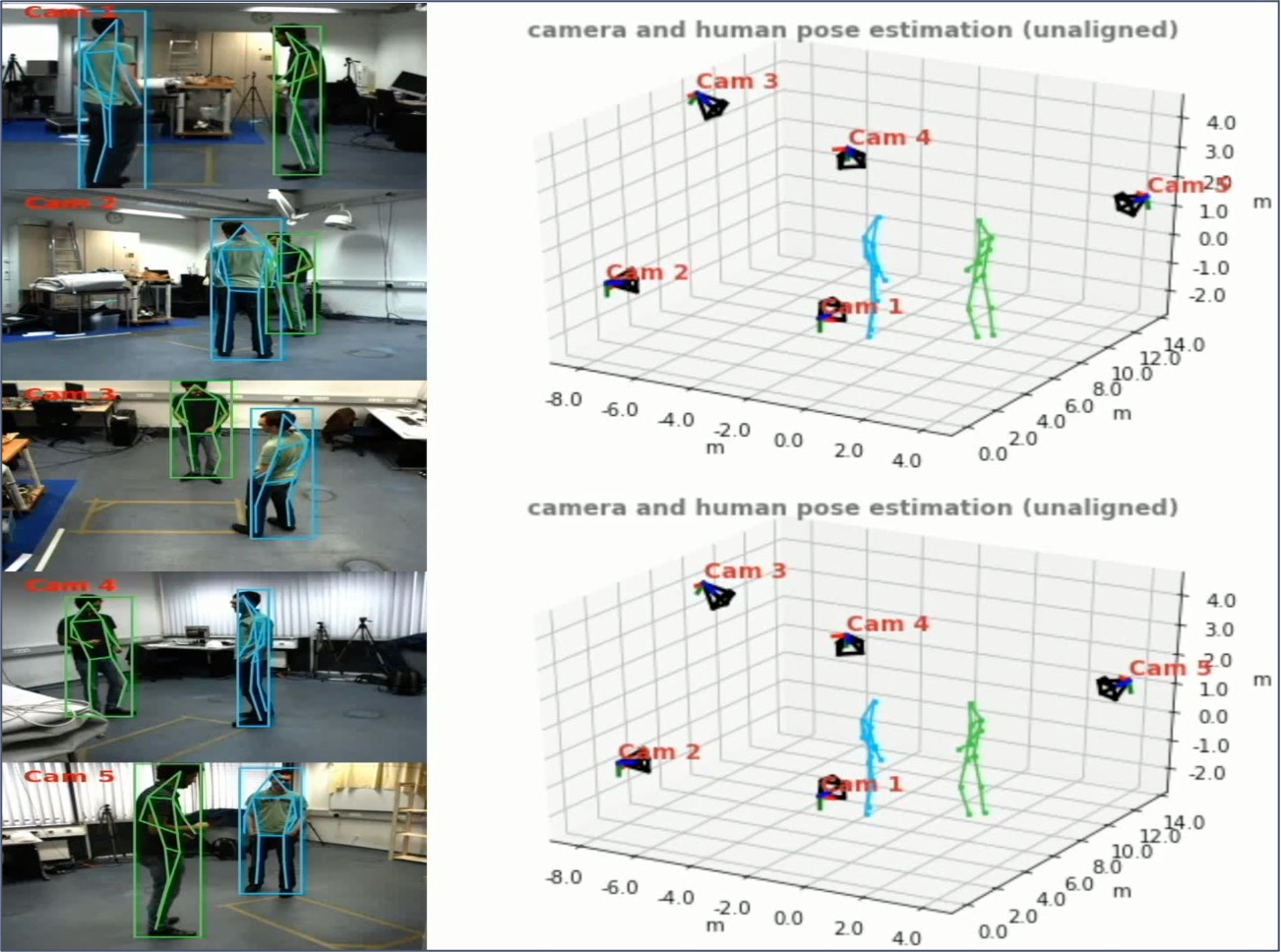}}
    & 
    \begin{itemize}
    \item Camera poses: Unknown
    \item Environment: Indoor
    \item Scene size: $\sim 12m\times 12m$
    \item \#Cameras: 5
    \item \#People: 2 - 3
    \end{itemize}
    & 
    \begin{itemize}
    \item \href{https://drive.google.com/file/d/1IH1zGlnsFqsOwY9vlgrC20zfLk8Ot3d6/view?usp=sharing}{\textcolor{teal}{Video\_Result\_Shelf}}
    \end{itemize} \\
    \bottomrule
\end{tabular}
\caption{Large-scale video results of human and camera pose estimation on a number of video sequences.  Our method does not uses camera poses for all data and generalizes well across a variant of scene sizes, indoor/outdoor environments, number of cameras, and number of people.  Please consider using the video links to check more visual results.}
\label{tab:pose_est_video}
\end{center}
\end{table*}
\begin{figure*}
\begin{center}
\includegraphics*[width=\linewidth]{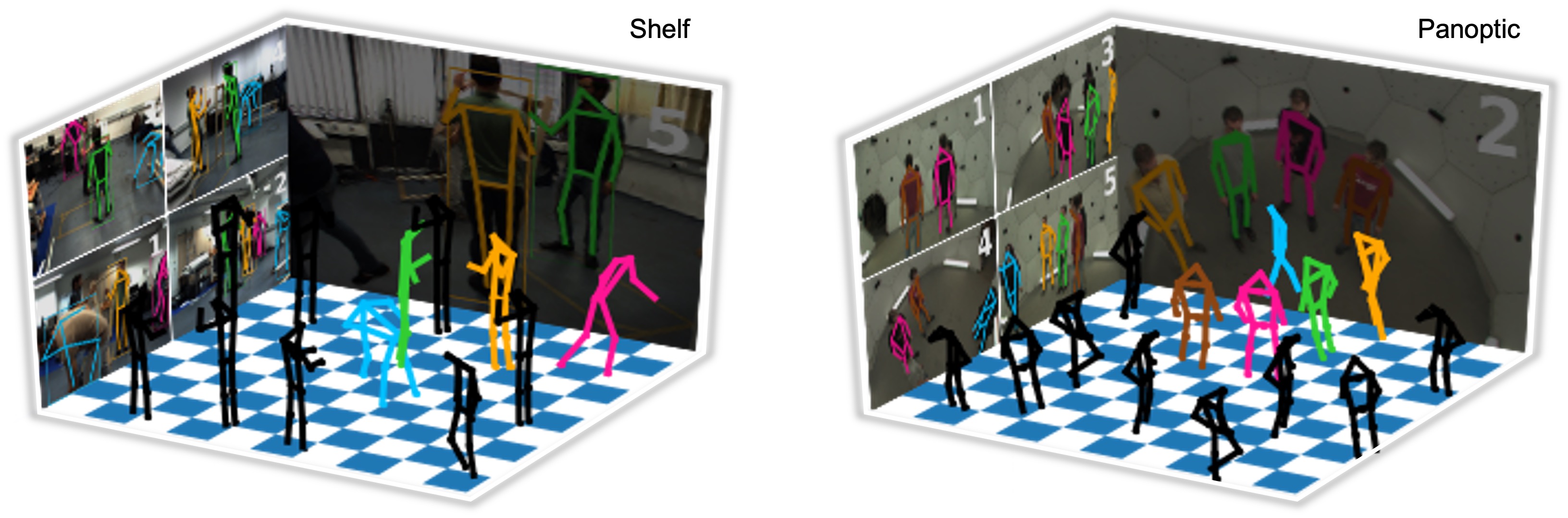}
\end{center}
\caption{Visual augmentation using our method with only multi-view images as input. Our method estimates camera poses, so we can generate realistic and geometrically plausible 3D visual augmentations through perspective projection.  Colored people are real, whose 3D poses are from estimation.  Black people are fake, inserted into the scene using the estimated camera poses.}
\label{fig:ar}
\end{figure*}
\section{Appendix}

\paragraph{Large-scale video results.}

For a more intuitive evaluation of our method, we present in Tab.\ref{tab:pose_est_video} 3D human and camera pose estimation results of our method on several datasets with different environment settings. The results include three open datasets (Panoptic, Shelf, and Campus) and two indoor and outdoor datasets we collected using arbitrarily set cameras (Office and Square).  In Tab.\ref{tab:pose_est_video}, the left column shows the results of cross-view people matching and 3D camera and human pose estimation, the middle column shows experiment setting, and the right column shows links to video results.  In each snapshot image, people are matched by color across camera views and 2D-3D, the top 3D figure shows 3D pose estimation result, where the depth ambiguity is solved by assuming the length of a person's lower leg is 0.5m, and the bottom 3D figure shows 3D pose estimation result, where the depth ambiguity is solved by assuming the distance between a camera pair to be fixed.  For the Square dataset, the scene size is large and people are away from the cameras, we found that a small camera angle estimation error will cause a large camera location change.  It is not good for visual understanding showing the camera poses, so we only show human pose estimation for this dataset.

From Tab.\ref{tab:pose_est_video}, we observe that our method can generalize across various environment settings, including different scene sizes, indoor/outdoor lighting conditions, number of cameras, and number of people.  Existing methods require either camera poses or 3D training data, while our method requires neither of them.  Our method is probably one of the first that can generalize across environments of these many different factors.  We also found that the regular pre-trained 2D pose detector can well-satisfy the quality requirement of our method on 2D human poses.  The detector we use for Office and Square in our experiment is HigherHRNet~\cite{cheng2020higherhrnet}.  For the open datasets, we use their provided 2D poses after occlusion processing. Finally, we observe that our method performs stably with equal to or less than four people in the scene.  When there are over five people in a narrow space, such as the Panoptic dataset, the performance will be impacted.  How to deal with severe occlusion while keeping the generalization ability would be an interesting direction to explore.

\paragraph{Virtual reality demonstration.} We show in Fig.\ref{fig:ar} one potential application scenario of our method, virtual reality.  Fig.\ref{fig:ar} was generated using only multi-view image sequences as input.  We match people across camera views, then estimate 3D human poses and 6 DoF camera poses.  Given arbitrary 3D human poses that are kinesics-wise reasonable, we project the 3D human poses into the reconstructed 3D scene using perspective projection.  As Fig.\ref{fig:ar} shows, the obtained virtual reality is realistic and geometrically plausible.  Existing deep methods can estimate human poses after training on specific datasets, however, cannot build such realistic virtual reality.  Other multi-stage pose estimation methods require camera poses in order to properly function and also cannot be directly used in virtual reality applications.  Compared to them, our method does not require expensive prior knowledge on the scene, i.e., camera poses and 3D training data, to work for human pose estimation, and also is much easier to be directly grabbed and used for applications, such as virtual reality.

\clearpage
{\small
\bibliographystyle{ieee_fullname}
\bibliography{egbib}

\begin{thebibliography}{10}\itemsep=-1pt

\bibitem{arthur2006k}
David Arthur and Sergei Vassilvitskii.
\newblock k-means++: The advantages of careful seeding.
\newblock Technical report, Stanford, 2006.

\bibitem{belagiannis20143d}
Vasileios Belagiannis, Sikandar Amin, Mykhaylo Andriluka, Bernt Schiele, Nassir
  Navab, and Slobodan Ilic.
\newblock 3d pictorial structures for multiple human pose estimation.
\newblock In {\em Proceedings of the IEEE Conference on Computer Vision and
  Pattern Recognition}, pages 1669--1676, 2014.

\bibitem{belagiannis20153d}
Vasileios Belagiannis, Sikandar Amin, Mykhaylo Andriluka, Bernt Schiele, Nassir
  Navab, and Slobodan Ilic.
\newblock 3d pictorial structures revisited: Multiple human pose estimation.
\newblock {\em IEEE transactions on pattern analysis and machine intelligence},
  38(10):1929--1942, 2015.

\bibitem{belagiannis2014multiple}
Vasileios Belagiannis, Xinchao Wang, Bernt Schiele, Pascal Fua, Slobodan Ilic,
  and Nassir Navab.
\newblock Multiple human pose estimation with temporally consistent 3d
  pictorial structures.
\newblock In {\em European Conference on Computer Vision}, pages 742--754.
  Springer, 2014.

\bibitem{bradley2000constrained}
Paul~S Bradley, Kristin~P Bennett, and Ayhan Demiriz.
\newblock Constrained k-means clustering.
\newblock {\em Microsoft Research, Redmond}, 20(0):0, 2000.

\bibitem{burenius20133d}
Magnus Burenius, Josephine Sullivan, and Stefan Carlsson.
\newblock 3d pictorial structures for multiple view articulated pose
  estimation.
\newblock In {\em Proceedings of the IEEE conference on computer vision and
  pattern recognition}, pages 3618--3625, 2013.

\bibitem{cao2017realtime}
Zhe Cao, Tomas Simon, Shih-En Wei, and Yaser Sheikh.
\newblock Realtime multi-person 2d pose estimation using part affinity fields.
\newblock In {\em Proceedings of the IEEE conference on computer vision and
  pattern recognition}, pages 7291--7299, 2017.

\bibitem{cheng2020higherhrnet}
Bowen Cheng, Bin Xiao, Jingdong Wang, Honghui Shi, Thomas~S Huang, and Lei
  Zhang.
\newblock Higherhrnet: Scale-aware representation learning for bottom-up human
  pose estimation.
\newblock In {\em Proceedings of the IEEE/CVF conference on computer vision and
  pattern recognition}, pages 5386--5395, 2020.

\bibitem{dong2019fast}
Junting Dong, Wen Jiang, Qixing Huang, Hujun Bao, and Xiaowei Zhou.
\newblock Fast and robust multi-person 3d pose estimation from multiple views.
\newblock In {\em Proceedings of the IEEE/CVF Conference on Computer Vision and
  Pattern Recognition}, pages 7792--7801, 2019.

\bibitem{ershadi2018multiple}
Sara Ershadi-Nasab, Erfan Noury, Shohreh Kasaei, and Esmaeil Sanaei.
\newblock Multiple human 3d pose estimation from multiview images.
\newblock {\em Multimedia Tools and Applications}, 77(12):15573--15601, 2018.

\bibitem{he2010laplacian}
Xiaofei He, Deng Cai, Yuanlong Shao, Hujun Bao, and Jiawei Han.
\newblock Laplacian regularized gaussian mixture model for data clustering.
\newblock {\em IEEE Transactions on Knowledge and Data Engineering},
  23(9):1406--1418, 2010.

\bibitem{huang2021dynamic}
Buzhen Huang, Yuan Shu, Tianshu Zhang, and Yangang Wang.
\newblock Dynamic multi-person mesh recovery from uncalibrated multi-view
  cameras.
\newblock In {\em 2021 International Conference on 3D Vision (3DV)}, pages
  710--720. IEEE, 2021.

\bibitem{huang2020end}
Congzhentao Huang, Shuai Jiang, Yang Li, Ziyue Zhang, Jason Traish, Chen Deng,
  Sam Ferguson, and Richard Yi~Da Xu.
\newblock End-to-end dynamic matching network for multi-view multi-person 3d
  pose estimation.
\newblock In {\em European Conference on Computer Vision}, pages 477--493.
  Springer, 2020.

\bibitem{joo2015panoptic}
Hanbyul Joo, Hao Liu, Lei Tan, Lin Gui, Bart Nabbe, Iain Matthews, Takeo
  Kanade, Shohei Nobuhara, and Yaser Sheikh.
\newblock Panoptic studio: A massively multiview system for social motion
  capture.
\newblock In {\em Proceedings of the IEEE International Conference on Computer
  Vision}, pages 3334--3342, 2015.

\bibitem{kalayeh2018human}
Mahdi~M Kalayeh, Emrah Basaran, Muhittin G{\"o}kmen, Mustafa~E Kamasak, and
  Mubarak Shah.
\newblock Human semantic parsing for person re-identification.
\newblock In {\em Proceedings of the IEEE conference on computer vision and
  pattern recognition}, pages 1062--1071, 2018.

\bibitem{kanungo2002efficient}
Tapas Kanungo, David~M Mount, Nathan~S Netanyahu, Christine~D Piatko, Ruth
  Silverman, and Angela~Y Wu.
\newblock An efficient k-means clustering algorithm: Analysis and
  implementation.
\newblock {\em IEEE transactions on pattern analysis and machine intelligence},
  24(7):881--892, 2002.

\bibitem{kuhn1955hungarian}
Harold~W Kuhn.
\newblock The hungarian method for the assignment problem.
\newblock {\em Naval research logistics quarterly}, 2(1-2):83--97, 1955.

\bibitem{lecun1998gradient}
Yann LeCun, L{\'e}on Bottou, Yoshua Bengio, and Patrick Haffner.
\newblock Gradient-based learning applied to document recognition.
\newblock {\em Proceedings of the IEEE}, 86(11):2278--2324, 1998.

\bibitem{li2019rethinking}
Wenbo Li, Zhicheng Wang, Binyi Yin, Qixiang Peng, Yuming Du, Tianzi Xiao, Gang
  Yu, Hongtao Lu, Yichen Wei, and Jian Sun.
\newblock Rethinking on multi-stage networks for human pose estimation.
\newblock {\em arXiv preprint arXiv:1901.00148}, 2019.

\bibitem{li2018harmonious}
Wei Li, Xiatian Zhu, and Shaogang Gong.
\newblock Harmonious attention network for person re-identification.
\newblock In {\em Proceedings of the IEEE conference on computer vision and
  pattern recognition}, pages 2285--2294, 2018.

\bibitem{liu2018pose}
Jinxian Liu, Bingbing Ni, Yichao Yan, Peng Zhou, Shuo Cheng, and Jianguo Hu.
\newblock Pose transferrable person re-identification.
\newblock In {\em Proceedings of the IEEE conference on computer vision and
  pattern recognition}, pages 4099--4108, 2018.

\bibitem{luo2019bag}
Hao Luo, Youzhi Gu, Xingyu Liao, Shenqi Lai, and Wei Jiang.
\newblock Bag of tricks and a strong baseline for deep person
  re-identification.
\newblock In {\em Proceedings of the IEEE/CVF conference on computer vision and
  pattern recognition workshops}, pages 0--0, 2019.

\bibitem{ma2022virtual}
Wei-Chiu Ma, Anqi~Joyce Yang, Shenlong Wang, Raquel Urtasun, and Antonio
  Torralba.
\newblock Virtual correspondence: Humans as a cue for extreme-view geometry.
\newblock In {\em Proceedings of the IEEE/CVF Conference on Computer Vision and
  Pattern Recognition}, pages 15924--15934, 2022.

\bibitem{moon1996expectation}
Todd~K Moon.
\newblock The expectation-maximization algorithm.
\newblock {\em IEEE Signal processing magazine}, 13(6):47--60, 1996.

\bibitem{mullner2011modern}
Daniel M{\"u}llner.
\newblock Modern hierarchical, agglomerative clustering algorithms.
\newblock {\em arXiv preprint arXiv:1109.2378}, 2011.

\bibitem{ng2001spectral}
Andrew Ng, Michael Jordan, and Yair Weiss.
\newblock On spectral clustering: Analysis and an algorithm.
\newblock {\em Advances in neural information processing systems}, 14, 2001.

\bibitem{palacio2019evaluation}
Julio-Omar Palacio-Ni{\~n}o and Fernando Berzal.
\newblock Evaluation metrics for unsupervised learning algorithms.
\newblock {\em arXiv preprint arXiv:1905.05667}, 2019.

\bibitem{perez2022matching}
Alejandro Perez-Yus and Antonio Agudo.
\newblock Matching and recovering 3d people from multiple views.
\newblock In {\em Proceedings of the IEEE/CVF Winter Conference on Applications
  of Computer Vision}, pages 3622--3631, 2022.

\bibitem{reddy2021tessetrack}
N~Dinesh Reddy, Laurent Guigues, Leonid Pishchulin, Jayan Eledath, and
  Srinivasa~G Narasimhan.
\newblock Tessetrack: End-to-end learnable multi-person articulated 3d pose
  tracking.
\newblock In {\em Proceedings of the IEEE/CVF Conference on Computer Vision and
  Pattern Recognition}, pages 15190--15200, 2021.

\bibitem{scarselli2008graph}
Franco Scarselli, Marco Gori, Ah~Chung Tsoi, Markus Hagenbuchner, and Gabriele
  Monfardini.
\newblock The graph neural network model.
\newblock {\em IEEE transactions on neural networks}, 20(1):61--80, 2008.

\bibitem{schubert2017dbscan}
Erich Schubert, J{\"o}rg Sander, Martin Ester, Hans~Peter Kriegel, and Xiaowei
  Xu.
\newblock Dbscan revisited, revisited: why and how you should (still) use
  dbscan.
\newblock {\em ACM Transactions on Database Systems (TODS)}, 42(3):1--21, 2017.

\bibitem{si2018dual}
Jianlou Si, Honggang Zhang, Chun-Guang Li, Jason Kuen, Xiangfei Kong, Alex~C
  Kot, and Gang Wang.
\newblock Dual attention matching network for context-aware feature sequence
  based person re-identification.
\newblock In {\em Proceedings of the IEEE conference on computer vision and
  pattern recognition}, pages 5363--5372, 2018.

\bibitem{solera2016tracking}
Francesco Solera, Simone Calderara, Ergys Ristani, Carlo Tomasi, and Rita
  Cucchiara.
\newblock Tracking social groups within and across cameras.
\newblock {\em IEEE Transactions on Circuits and Systems for Video Technology},
  27(3):441--453, 2016.

\bibitem{song2018mask}
Chunfeng Song, Yan Huang, Wanli Ouyang, and Liang Wang.
\newblock Mask-guided contrastive attention model for person re-identification.
\newblock In {\em Proceedings of the IEEE conference on computer vision and
  pattern recognition}, pages 1179--1188, 2018.

\bibitem{suh2018part}
Yumin Suh, Jingdong Wang, Siyu Tang, Tao Mei, and Kyoung~Mu Lee.
\newblock Part-aligned bilinear representations for person re-identification.
\newblock In {\em Proceedings of the European conference on computer vision
  (ECCV)}, pages 402--419, 2018.

\bibitem{sun2019deep}
Ke Sun, Bin Xiao, Dong Liu, and Jingdong Wang.
\newblock Deep high-resolution representation learning for human pose
  estimation.
\newblock In {\em Proceedings of the IEEE/CVF Conference on Computer Vision and
  Pattern Recognition}, pages 5693--5703, 2019.

\bibitem{triggs1999bundle}
Bill Triggs, Philip~F McLauchlan, Richard~I Hartley, and Andrew~W Fitzgibbon.
\newblock Bundle adjustment—a modern synthesis.
\newblock In {\em International workshop on vision algorithms}, pages 298--372.
  Springer, 1999.

\bibitem{tu2020voxelpose}
Hanyue Tu, Chunyu Wang, and Wenjun Zeng.
\newblock Voxelpose: Towards multi-camera 3d human pose estimation in wild
  environment.
\newblock In {\em European Conference on Computer Vision}, pages 197--212.
  Springer, 2020.

\bibitem{vaswani2017attention}
Ashish Vaswani, Noam Shazeer, Niki Parmar, Jakob Uszkoreit, Llion Jones,
  Aidan~N Gomez, {\L}ukasz Kaiser, and Illia Polosukhin.
\newblock Attention is all you need.
\newblock {\em Advances in neural information processing systems}, 30, 2017.

\bibitem{wagstaff2001constrained}
Kiri Wagstaff, Claire Cardie, Seth Rogers, Stefan Schr{\"o}dl, et~al.
\newblock Constrained k-means clustering with background knowledge.
\newblock In {\em Icml}, volume~1, pages 577--584, 2001.

\bibitem{wu2021graph}
Size Wu, Sheng Jin, Wentao Liu, Lei Bai, Chen Qian, Dong Liu, and Wanli Ouyang.
\newblock Graph-based 3d multi-person pose estimation using multi-view images.
\newblock In {\em Proceedings of the IEEE/CVF International Conference on
  Computer Vision}, pages 11148--11157, 2021.

\bibitem{xu2021wide}
Yan Xu, Yu-Jhe Li, Xinshuo Weng, and Kris Kitani.
\newblock Wide-baseline multi-camera calibration using person
  re-identification.
\newblock In {\em Proceedings of the IEEE/CVF Conference on Computer Vision and
  Pattern Recognition}, pages 13134--13143, 2021.

\bibitem{zhang2021direct}
Jianfeng Zhang, Yujun Cai, Shuicheng Yan, Jiashi Feng, et~al.
\newblock Direct multi-view multi-person 3d pose estimation.
\newblock {\em Advances in Neural Information Processing Systems}, 34, 2021.

\bibitem{zhang2022voxeltrack}
Yifu Zhang, Chunyu Wang, Xinggang Wang, Wenyu Liu, and Wenjun Zeng.
\newblock Voxeltrack: Multi-person 3d human pose estimation and tracking in the
  wild.
\newblock {\em IEEE Transactions on Pattern Analysis and Machine Intelligence},
  2022.

\bibitem{zheng2015scalable}
Liang Zheng, Liyue Shen, Lu Tian, Shengjin Wang, Jingdong Wang, and Qi Tian.
\newblock Scalable person re-identification: A benchmark.
\newblock In {\em Proceedings of the IEEE international conference on computer
  vision}, pages 1116--1124, 2015.

\bibitem{zhong2018camera}
Zhun Zhong, Liang Zheng, Zhedong Zheng, Shaozi Li, and Yi Yang.
\newblock Camera style adaptation for person re-identification.
\newblock In {\em Proceedings of the IEEE conference on computer vision and
  pattern recognition}, pages 5157--5166, 2018.

\end{thebibliography}
}

\end{document}